\documentclass[Afour,sageh,times]{sagej}

\input{config.sty}
\input{glossary.sty}

\def\volumeyear{2026}
\def\journalname{Submitted to: The International Journal of Robotics Research}
\runninghead{Hou \textit{et~al.}}

\begin{document}

\title{Linear Temporal Logic Translation via Human-Inspired Self-Constrained Reasoning for Robot Task Specification}

\author{%
Haofei Hou$^1$\thanks{These authors contributed equally to this work.},
Fanxu Meng$^1$\footnotemark[1],
Shunyi Zhao$^2$\footnotemark[1],
Kairui Yang$^1$,
Mengchen Cai$^1$,\\
Lecheng Ruan$^1$\thanks{Corresponding author.},
Qining Wang$^1$\footnotemark[2]
}

\affiliation{%
$^*$ These authors contributed equally to this work.\\
$^1$ School of Advanced Manufacturing and Robotics, Peking University, Beijing 100871, China.\\
$^2$ School of Integrated Circuits, Peking University, Beijing 100871, China.
}

\corrauth{%
Lecheng Ruan, Email: \href{mailto:ruanlecheng@ucla.edu}{ruanlecheng@ucla.edu}\\[0.5em]
Qining Wang, Email: \href{mailto:qiningwang@pku.edu.cn}{qiningwang@pku.edu.cn}
}

\begin{abstract}
Many robotic tasks are temporally extended and demand precise specifications of subgoals, constraints, and their temporal ordering. Yet human operators typically communicate such tasks in natural language, which is inherently ambiguous, underspecified, and context dependent. Translating human instructions into formal task specifications, such as \acf{ltl}, is therefore essential for verifiable and safe robotic execution. Existing \acs{llm}-based translators attempt to bridge this gap through open-ended reasoning or post-hoc constraint enforcement, but the former may violate domain constraints, whereas the latter can disrupt the reasoning needed for novel instructions. This paper proposes \ac{scr}, a framework that mitigates this trade-off by internalizing structural knowledge into the model's decision-making process rather than imposing it as an external filter. By combining a structural constraint representation with a hierarchical decision-making formulation, \ac{scr} guides reasoning within a formally grounded space while preserving adaptability to unseen instructions. Experiments show that \ac{scr} improves both domain-constraint satisfaction and generalization, providing an effective and interpretable approach for translating human intent into verifiable specifications for robotic execution.
\end{abstract}

\maketitle

\section{Introduction}

As physical extensions of human capabilities, robots are designed to execute complex human intentions in the real world~\citep{intelligence2025pi_, ma2026survey}. A wide range of tasks delegated to robots by humans are temporally extended, unfolding over multiple interdependent stages rather than being defined by a single objective~\citep{liu2023lang2ltl}. Examples include cooking assistance, urban navigation, and robotic assembly, which involve sequences of sub-tasks governed by ordering constraints and time-varying requirements~\citep{mavrogiannis2024cook2ltl, liu2024lang2ltl, mendoza2024translating, rabiei2025ltlcodegen, brodo2026property}. Prior work has demonstrated that, without precise specifications, even minor misinterpretations can lead to incorrect or unsafe behavior~\citep{pan2023data, guo2026one}. Consequently, reliable execution requires unambiguous definitions of both the goals and their temporal dependencies.

\ac{nl} provides an intuitive interface for humans to communicate task intentions, yet it is inherently ambiguous, underspecified, and dependent on context~\citep{ahn2022can}. Robotic systems, however, require these intentions to be grounded in representations that are precise, task-relevant, and executable~\citep{shi2024autodsl, wang2026logicflow}. This gap is especially prominent in \ac{hri}, where users in voice-commanded assistance, domestic service, and laboratory automation often describe temporally extended tasks in everyday language, while the robot must infer the corresponding goals, constraints, and ordering relations~\citep{tellex2011understanding, rankin2021robotic, argenziano2025defining}. Asking users to provide these specifications directly in formal logic is not a practical alternative, since it requires expertise in symbolic reasoning and familiarity with logical syntax that most end users lack~\citep{pakonen2016user, santos2025updating, schlor2006using, ma2025bridging}. Bridging this gap by automatically translating \ac{nl} instructions into formal task specifications has therefore become a central challenge at the intersection of robotics, formal methods, and natural language understanding~\citep{chen2023nl2tl, germiniani2025systematic}.

\begin{figure}[t]
    \centering
    \includegraphics[width=0.95\linewidth]{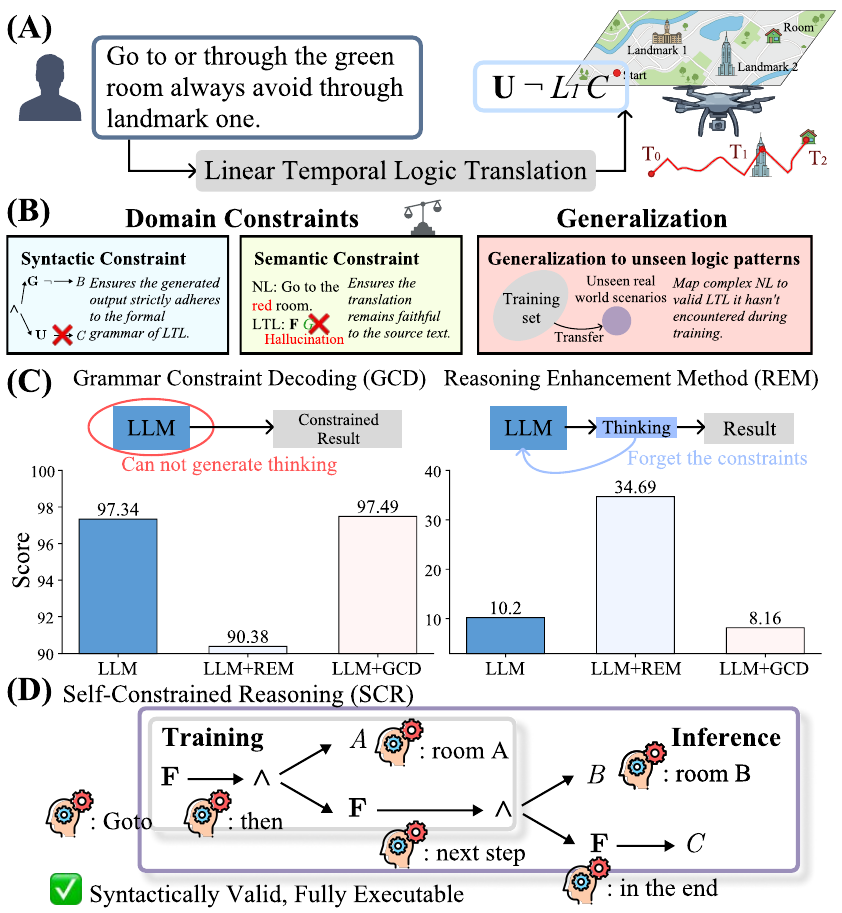}
    \caption{\textbf{Challenges and methodologies in linear temporal logic translation.} \textbf{(A)} A demonstration of translating a natural language instruction into a formal \ac{ltl} expression to guide autonomous systems. \textbf{(B)} Challenges in \ac{ltl} translation: domain constraints and generalization. \textbf{(C)} Comparative performance and pitfalls of two mainstream methods for existing challenges. Experiments demonstrate that while constraint-based methods improve adherence to domain constraints, they often fail to generate proper reasoning, leading to a decrease in the model's generalization ability. Conversely, reasoning-based methods enhance generalization capabilities but tend to overlook specific domain constraints during the process. \textbf{(D)} Our approach effectively balances domain-specific constraints with generalization capabilities, demonstrating robust performance even when presented with novel and increasingly complex task specifications. Furthermore, the method ensures strict adherence to grammatical rules to guarantee executability, while enforcing textual alignment between the output and human intent.}
    \label{fig:problem}
\end{figure}

\acf{ltl} offers a structured and expressive formalism for specifying temporally extended tasks, and has therefore become a widely used target representation for translating \ac{nl} instructions into formal specifications, as illustrated in Fig.~\ref{fig:problem}\textbf{(A)}. By extending propositional logic with temporal operators, \ac{ltl} can express properties that must hold over complete execution trajectories rather than at individual states~\citep{raman2014model, wang2025conformalnl2ltl}. For example, an instruction such as ``go to room one, then to room three, and always avoid landmark one'' can be encoded as a single specification that combines an ordering requirement with a safety constraint. Given such a specification, correct-by-construction controllers or motion plans can be synthesized to provably satisfy the desired constraints~\citep{gopalan2018sequence, patel2020grounding, gundana2022event, liu2026zero}. This connection between high-level task specification, formal verification, and controller synthesis makes \ac{ltl} particularly attractive for safety-critical and long-horizon robotic applications~\citep{wu2025selp}.

Translating \ac{nl} instructions into \ac{ltl} presents unique challenges compared to conventional language-to-structure mapping~\citep{gopalan2018sequence, oh2019planning, quartey2025verifiably}, as illustrated in Fig.~\ref{fig:problem}\textbf{(B)}. On one hand, specification tasks impose strict domain constraints, including both syntactic and semantic requirements. Generated formulas must adhere to the operator grammar of \ac{ltl}~\citep{baier2008principles, li2026environment}, respect domain-specific action semantics, and remain executable within robotic environments~\citep{shi2024autodsl}. On the other hand, robots frequently encounter out-of-distribution instructions and novel task compositions, requiring strong generalization beyond the training distribution~\citep{liu2024lang2ltl, guo2025castl}. Effectively addressing these demands requires methods that enforce strict domain conformance while generalizing to novel instructions, making this a fundamental challenge in translating \ac{nl} into \ac{ltl}.

Existing methods have approached this problem from two broad directions. Constraint-guided decoding methods prioritize formal correctness by restricting the output space during inference~\citep{hokamp2017lexically, kumar-etal-2022-gradient}. Such constraints are effective in preventing syntactically invalid formulas, but overly rigid enforcement can interfere with the model’s intermediate reasoning process, which is often important for interpreting complex or unfamiliar instructions~\citep{beurer2024guiding, english2025grammar}. In contrast, reasoning-enhanced methods improve generalization by encouraging the model to generate intermediate logical steps through strategies such as \ac{cot}, Self-Consistency, and Tree-of-Thought~\citep{wei2022chain, wang2022self, yao2023tree, guo2025deepseek}. However, because these methods rely on open-ended generation, they remain susceptible to hallucinations and may overlook domain-specific constraints~\citep{yao2025reasoning, sun2025detection}. As illustrated in Fig.~\ref{fig:problem}\textbf{(C)}, these trends reveal a persistent trade-off: stronger passive constraint enforcement tends to reduce reasoning-based generalization, whereas greater reasoning flexibility can weaken constraint adherence.

Recent work has sought to mitigate this trade-off through hybrid mechanisms that combine open-ended reasoning with symbolic structure~\citep{liu2024we, quansah2026neuronl2ltl}. The reason-then-constrain paradigm, exemplified by In-Writing, decouples exploratory reasoning from structured formatting, allowing models to reason in natural language before invoking constraint enforcement~\citep{nguyen2026thinking}. Grammar augmentation methods, such as CRANE, introduce additional primitives into the reasoning process to help determine when grammatical constraints should be applied~\citep{banerjee2025crane, agarwal2025think}. Multi-stage pipelines, such as DICORE, use an open-ended \ac{llm} for broad task discovery and an FSM-based \ac{llm} for structural mapping~\citep{parekh2025dicore}. Together, these studies suggest that progress in \ac{nl}-to-\ac{ltl} translation depends not merely on stronger filtering, but on neuro-symbolic mechanisms that preserve reasoning flexibility while maintaining formal discipline.

Although these methods partially mitigate the tension between constraint enforcement and reasoning~\citep{nguyen2026thinking}, they generally rely on post-hoc correction or external intervention rather than integrating constraints throughout the reasoning process. This reliance reflects two pervasive, underlying assumptions, which ultimately constrain current approaches. First, constraints are typically treated as external filters applied during inference, rather than as formal knowledge internalized during training~\citep{dash2022review, liu2026hard}. While such filters can intercept syntactic errors, they are less effective at resolving semantic inconsistencies: a forcibly corrected formula may satisfy the grammar of \ac{ltl} yet deviate from the user's intended task~\citep{liu2024we, loula2025syntactic}. Second, reasoning is often modeled as a linear token-generation process, which makes it difficult to regulate intermediate logic without globally restricting the model's expressive capacity~\citep{wei2022chain, zhou2026policy}. This is problematic for robotic task execution, where procedural safety requires intermediate inference steps to remain verifiable and consistent with task constraints~\citep{guo2025deepseek, ildizlearning}. When latent reasoning proceeds without such regulation, the model may neglect constraints embedded in the input and ultimately produce semantically inconsistent or unsafe actions~\citep{xu2025toward}. Together, these limitations suggest that existing methods address the symptoms of the constraint--reasoning conflict rather than its underlying cause.

Insights into the limitations of externalized constraints and linear reasoning can be drawn from human language processing. Linguistic theory, particularly Chomsky's Universal Grammar, posits that language production is guided by internalized grammatical structures~\citep{chomsky2014aspects, verkerk2026enduring}, and psycholinguistic and neuroscientific evidence suggests that comprehension and reasoning operate over hierarchical representations rather than purely linear sequences~\citep{rayner1975perceptual, chomsky2000new,  keshishian2026parallel}. These perspectives suggest a useful analogy for specification learning: constraints need not be applied only after reasoning, but can instead shape the reasoning process itself.

Inspired by these observations, we introduce \acf{scr}, a framework that formulates \ac{nl}-to-\ac{ltl} translation as a cognitively bounded process. Our framework reflects a mechanism where structural constraints actively govern hierarchical reasoning, much like how internalized grammar inherently bounds human language. This framework is driven by two core principles: 1. Structural knowledge is internalized during the training process, rather than applying constraints as external post-hoc filters. This forces the model to operate within a valid constraint space, eliminating syntactic errors at the source. 2. Specification generation is treated as a hierarchical planning procedure, where each intermediate reasoning step is anchored by domain-valid constructs. This allows the model to explore novel linguistic expressions while maintaining formal correctness. Built upon these two principles, the \acf{scr} design directly addresses the limitations of prior methods. By internalizing constraints and enabling verifiable intermediate reasoning, our approach reduces syntactic and semantic errors when mapping unseen human intent to verifiable robotic specifications, as illustrated in Fig.~\ref{fig:problem}\textbf{(D)}.

Our contributions are summarized as follows:
\begin{enumerate}
    \item We propose \acf{scr}, a cognitively inspired framework for \ac{nl}-to-\ac{ltl} translation that generalizes to novel instructions while maintaining strict domain-constraint satisfaction.
    \item We tailor the \ac{scfg} specifically to the \ac{ltl} translation task and introduce an \ac{ltl}-\acs{scfg} constraint representation. This internalizes domain knowledge during training and decision-making, allowing the model to reason over new information while remaining within a valid constraint space.
    \item We develop a hierarchical decision-making formulation for \ac{ltl} translation that enables hierarchical reasoning and constraint injection throughout generation.
    \item We conduct extensive experiments showing that \ac{scr} consistently outperforms constrained and unconstrained baselines in domain-constraint satisfaction, generalization, safety violation reduction, and interpretability.
\end{enumerate}

\section{Methodology}

\subsection{Preliminaries}

In this paper, robotic tasks are specified using \ac{ltl}, which augments propositional logic with temporal operators to express properties of trajectories over time. We consider the planning problem for discrete-time systems governed by the dynamics $x_{t+1} = f(x_t, u_t)$, where $x \in \mathcal{X}$ represents the state and $u \in \mathcal{U}$ the control input. The control objective is to synthesize a policy that satisfies a high-level task specification derived from a \ac{nl} instruction $s$.

The \ac{ltl} specifications considered in this work follow a recursive grammar in prefix notation:
\begin{equation}
    \varphi ::= p 
    \mid \neg \varphi_1 
    \mid \land \varphi_1 \varphi_2 
    \mid \lor \varphi_1 \varphi_2 
    \mid \mathbf{G}\varphi 
    \mid \mathbf{F}\varphi 
    \mid \mathbf{U} \varphi_1 \varphi_2, 
    \label{equ:ltl-grammar}
\end{equation}
where $p \in \mathcal{P} = \{p_i\}_{i=1}^{N}$ are \acp{ap}. 
In this work, the \acp{ap} correspond to sets of salient regions in the state space that the robot may visit or avoid. Within this framework, each \ac{ap} is associated with a constrained region in the state space defined as $x \models p_i \iff g_i(x) \le 0$, where $g_i: \mathcal{X} \to \mathbb{R}$ represents a constraint function. The temporal operators are defined as follows: $\mathbf{G}\phi$ denotes that the condition $\phi$ must hold \textbf{globally} for all time; $\mathbf{F}\phi$ denotes that $\phi$ must hold \textbf{eventually} (\ie, there exists some time step $t$ where $\phi$ is true); $\mathbf{U} \phi_1 \phi_2$ denotes that $\phi_1$ must hold for all time steps \textbf{until} $\phi_2$ holds. 

\subsection{Cognitive Foundations and Overview of the Framework}

\begin{figure*}[t]
    \centering
    \includegraphics[width=\textwidth]{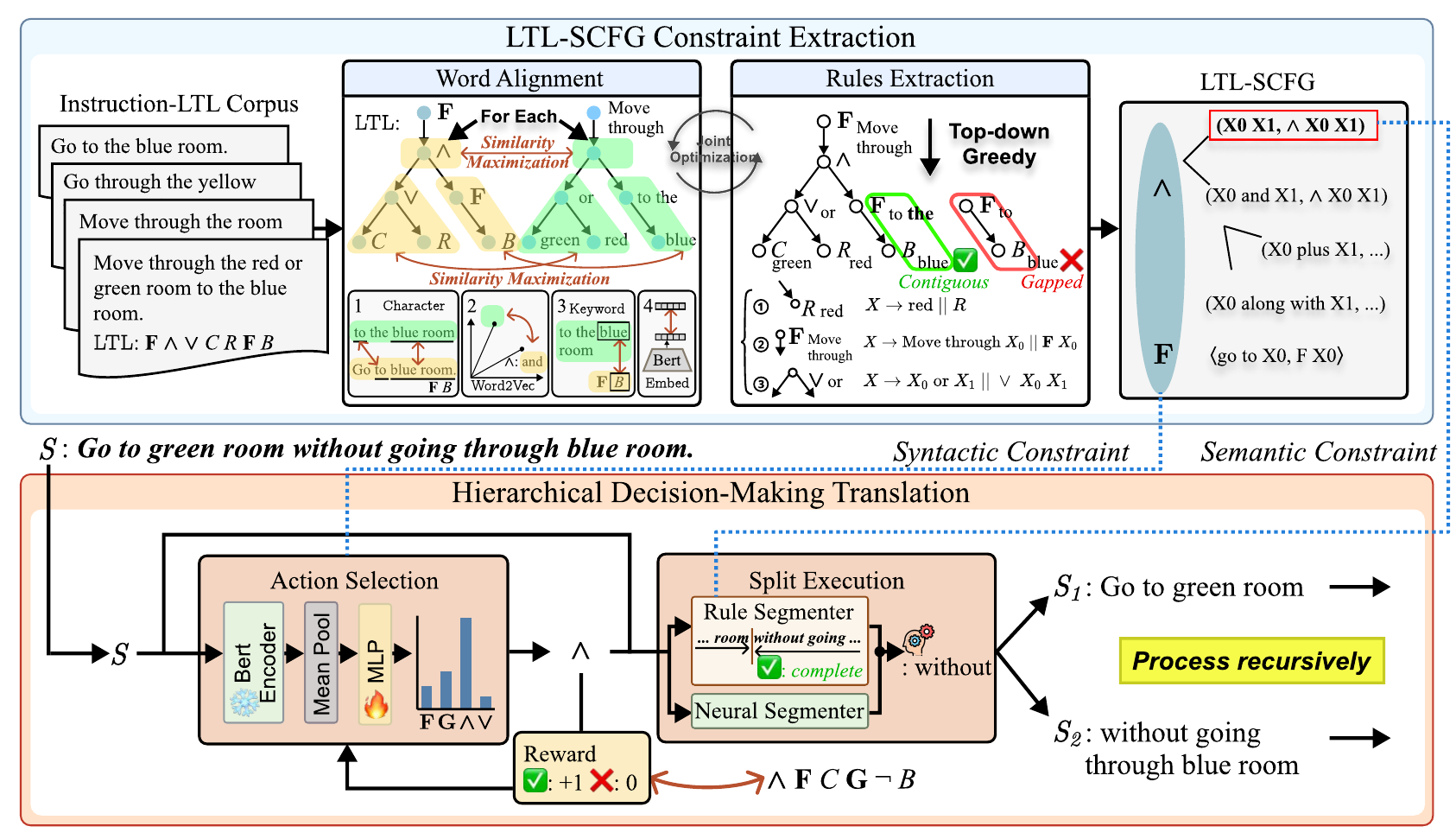}
    \caption{\textbf{Overview of the proposed framework.} The pipeline extracts an \ac{ltl}-\acs{scfg} from parallel corpora (top) and utilizes it to constrain the \ac{rl} agent's action space during hierarchical translation (bottom).}
    \label{fig:pipeline}
\end{figure*}

To resolve the inherent tension between flexible reasoning and rigid constraint satisfaction, we propose a hierarchical framework for translating \ac{nl} into \ac{ltl}. Our approach adopts an \ac{scr} architecture that internalizes domain knowledge directly into the model's decision-making dynamics.

Our framework is inspired by psycholinguistic evidence suggesting that human language production is a hierarchical process governed by internalized structural rules~\citep{chomsky2014aspects, frankland2020concepts}. When a human expert translates a complex instruction such as ``Visit room A, then B, while avoiding C,'' they do not generate symbols blindly; they intuitively identify \acp{ap} and decompose the temporal logic into a nested tree of subgoals~\citep{correa2023humans}.

As illustrated in Fig. \ref{fig:pipeline}, our framework achieves a synergy of flexible reasoning and rigid constraint satisfaction through a closed-loop hierarchical planning architecture. This ensures that the model's reasoning is naturally bounded by the formal grammar of \ac{ltl}, preventing the hallucinations common in unconstrained reasoning models while maintaining the fluidity often lost in post-hoc filtering methods.

The proposed framework bridges data-driven grammar learning with hierarchical execution. First, the framework ingests a parallel corpus of language instructions and \ac{ltl} formulas to automatically construct a synchronized \ac{ltl}-\acs{scfg} grammar. This is achieved through a word alignment that optimizes multi-dimensional semantic and lexical similarities to map formal operators to text spans, followed by a hierarchical rule extraction that employs a top-down greedy algorithm to recursively compose grammar rules along the \ac{ltl} syntax tree. Based on this learned structural space, the system then translates novel instructions by decoupling high-level logical reasoning from low-level text segmentation: an \ac{rl}-trained language agent policy dynamically selects the dominant outermost \ac{ltl} operator, while an adaptive segmenter simultaneously decomposes the remaining text into constituent sub-sentences. This decision-and-split process recursively descends along the \ac{ltl} syntax tree until all subtasks are resolved, ultimately outputting a verifiable, structurally sound formal \ac{ltl} formula.

Unlike standard \ac{rl} approaches that explore an unconstrained action space, our agent operates within a space defined by this prefix grammar:
\begin{itemize}
    \item \textbf{Syntactic Constraints:} The \ac{ltl} grammar serves as the primary decision-making objective, ensuring that every generated formula is syntactically valid and executable within the robot's domain.
    \item \textbf{Semantic Constraints:} The source language provides the semantic foundation, allowing the model to segment the instruction into manageable subtasks that correspond to specific logical operators and operands.
\end{itemize}

\subsection{LTL-SCFG Constraint Extraction}\label{subsec:LTL-SCFG}

In this section, we introduce the \ac{ltl}-\acs{scfg} and describe the procedure for extracting and constructing the constraint $G$ from a parallel corpus $C = \{(s_i,\varphi_i)\}$, where $s_i$ denotes an \ac{nl} instruction and $\varphi_i$ represents its corresponding \ac{ltl} formula. \ac{ltl}-\acs{scfg} allows the policy to incorporate and reason about new information while operating within a formal constraint space. The extraction process consists of two major stages: word alignment and rule extraction.

\paragraph{LTL-SCFG Definition}

The \acf{scfg} has a long-established role in statistical machine translation. Formally, an \ac{scfg} comprises a set of production rules $R$ of the following form:
\begin{equation}
    X \rightarrow \alpha \parallel \beta, 
\end{equation}
where $X$ is a nonterminal symbol, $\alpha \in (\{X\} \cup \Sigma_\alpha)^*$ is a sequence of terminal and/or nonterminal symbols in the source language (English), and $\beta \in (\{X\} \cup \Sigma_\beta)^*$ is the corresponding sequence in the target language (\ie, \ac{ltl}). The terminal symbol sets of the source and target languages are denoted as $\Sigma_\alpha$ and $\Sigma_\beta$, respectively. This parallel structure captures the synchronization between the source and target representations.

For example, consider a rule in prefix notation:
\begin{equation}
X \rightarrow X_0 \text{ and } X_1 \parallel \land \ X_0 \ X_1.
\end{equation}
An \ac{scfg} typically uses a single nonterminal symbol, where subscripts indicate the correspondence between source and target nonterminals. By recursively applying such rules, a complete translation can be generated in a top-down manner. However, since \acp{nl} generally do not conform to LL($k$) grammar constraints, top-down parsing is often infeasible for natural language. Consequently, traditional statistical translation models usually employ bottom-up parsing instead.

In the context of \ac{ltl} translation, the set of \acp{ap} $\mathcal{P}$ is comparatively small. When we consider \ac{ltl} expressions containing at most one operator and restrict operands to nonterminal symbols, the total number of valid target-side expressions is bounded by $|\mathcal{P}| + 6$. Owing to this property, humans naturally reason about \ac{ltl} expressions in a top-down manner---contrasting with the mixed strategies commonly observed in general \ac{nl} translation. This cognitive observation provides the theoretical foundation for imposing structural constraints on the language model.

We define \ac{ltl}-\ac{scfg} as a tuple $(X, \Sigma_\alpha, \Sigma_\beta, f)$, where $\Sigma_\beta = \mathcal{P} \cup \{ \neg, \land, \lor, \mathbf{G}, \mathbf{F}, \mathbf{U}\}$, and the mapping function $f$ associates each target-side operator $o \in \Sigma_\beta$ with its corresponding set of source-side realizations and weights $\{(\alpha_i, w_i) \mid \alpha_i \in (\{X\} \cup \Sigma_\alpha)^*, w_i > 0\}$. This formulation enables $\Sigma_\beta$ to serve as the decision-making target---specifically, determining the outermost operator or primitive of the current sentence---thereby constraining the exploration space of \ac{rl} during translation. Simultaneously, the source-side expressions provide the foundation for task decomposition.

\paragraph{Word Alignment} 

As defined in the \ac{ltl}-\ac{scfg} framework, identifying the textual counterparts of \acp{ap} and operators in \ac{ltl} is essential, as it directly determines the construction of the mapping function $f$. Specifically, word alignment, denoted as $\omega_i: \Sigma_\beta \rightarrow \Sigma_\alpha^*$, aims to establish a correspondence between each \ac{ap} or operator in an \ac{ltl} formula $\varphi_i$ and its corresponding textual element $s_{m,n} \in s_i$.

Although one could attempt to optimize the mapping by directly maximizing similarity, such local optimization is inherently flawed as it lacks global contextual consideration, often resulting in suboptimal alignments. Moreover, the existence of synonyms in \ac{nl} further increases the risk of incorrect alignments. Given the back-translation-friendly nature of \ac{ltl}, it is feasible to define an \ac{scfg} that translates \ac{ltl} into structured English~\citep{pan2023data}. Aligning the formulas within this structured English space facilitates clause-level reasoning and significantly reduces alignment complexity.

According to the definition in Eq. \eqref{equ:ltl-grammar}, the syntactic tree of any \ac{ltl} formula can be directly derived from its prefix form. Each \ac{ltl} formula $\varphi$ can thus be represented as a rooted, ordered syntactic tree $T_\varphi = (V_\varphi, E_\varphi, \lambda_\varphi, \ell_\varphi, \mathbf{r}_\varphi)$, where $V_\varphi$ denotes the set of nodes, $\mathbf{r}_\varphi \in V_\varphi$ is the root, $E_\varphi \subseteq V_\varphi \times V_\varphi$ is the set of directed edges, $\lambda_\varphi : V_\varphi \rightarrow \Sigma_\beta$ is a labeling function, and $\ell_\varphi : V_\varphi \rightarrow \Sigma_\beta^*$ denotes a sub-\ac{ltl} labeling function. The structure satisfies $\ell_\varphi(\mathbf{r}_\varphi) = \varphi$ and 
\begin{equation}\label{equ:build_tree}
\left\{
\begin{aligned}
\lambda_\varphi(v)=p,\ \ell_\varphi(v)=p_i
&\Leftrightarrow\ p_i \in \mathcal{P},\\
\lambda_\varphi(v)=\mathbf{o},\ \ell_\varphi(v)=\mathbf{o}\,\ell_\varphi(v_1)
&\Leftrightarrow\
\substack{(v,v_1)\in E_\varphi \\ \mathbf{o}\in\{\neg,\mathbf{G},\mathbf{F}\}},\\
\lambda_\varphi(v)=\mathbf{o},\ \ell_\varphi(v)=\mathbf{o}\,\ell_\varphi(v_1)\,\ell_\varphi(v_2)
&\Leftrightarrow\
\substack{(v,v_1)\in E_\varphi \\ (v,v_2)\in E_\varphi \\ \mathbf{o}\in\{\land,\lor,\mathbf{U}\}}.
\end{aligned}
\right.
\end{equation}
Since \ac{ltl} can be naturally represented as a syntax tree, we leverage \ac{llm}-based sampling to construct a plausible mapping function $f_B$, which is subsequently used to build the back-translation grammar $G_B$ for \ac{ltl}-\ac{scfg}. By recursively applying $f_B$, structured English sentences $s_B(\ell_\varphi(v))$ can be generated for any \ac{ltl} clause $\ell_\varphi(v\in V_\varphi)$ through an \ac{mcmc} process:
\begin{equation}
\begin{aligned}
    s_B(\ell_\varphi(v)) & \sim \text{MCMC}(v, f_B) \\
    &\sim P(s_B(\ell_\varphi(v)) \mid s_B(\ell_\varphi(v_s))) \quad (v,v_s) \in E_\varphi \\
    &\sim \text{Uniform}(f_B(\lambda_\varphi(v))) \times \textstyle \prod_{v_s} P(s_B(\ell_\varphi(v_s))),
\end{aligned}
\end{equation}
where $v_s$ denotes a child node of $v$ in the syntax tree $T_\varphi$, and $\text{Uniform}(L)$ denotes probability mass function of the discrete uniform distribution over the set $L$.

To address the limitations of direct similarity maximization, we formulate the following optimization objective:
\begin{equation}\label{eq:omega_opt}
\begin{aligned}
\omega_i
&= \arg\max_{\omega_i} J(\omega_i, s_i, T_{\varphi_i}) \\
&= \arg\max_{\omega_i} \textstyle\sum_{v \in V_{\varphi_i}}
\Big[
  S\big(\text{MCMC}(v,\omega_i), s_B(\ell_{\varphi_i}(v))\big) \\
&\hspace{3.2em} + S\big(\omega_i(\lambda_{\varphi_i}(v)), f_B(\lambda_{\varphi_i}(v))\big)
\Big].
\end{aligned}
\end{equation}
where
{\small\begin{equation}\label{equ:similarity}
\begin{aligned}
S(s_{m,n}, L)
&= \frac{1}{|L|}\sum_{\xi \in L}
\Bigl[
  \gamma_1 S_{\text{char}}(s_{m,n}, \xi)
 + \gamma_2 S_{\text{w2v}}(s_{m,n}, \xi) \\
&\hspace{4.2em}
 + \gamma_3 S_{\text{key}}(s_{m,n}, \xi)
 + \gamma_4 S_{\text{sent}}(s_{m,n}, \xi)
\Bigr].
\end{aligned}
\end{equation}}
where each term measures a specific aspect of similarity: $S_{\text{char}}$ for character-level matching, $S_{\text{w2v}}$ for word-level semantic similarity, $S_{\text{key}}$ for key-term overlap, and $S_{\text{sent}}$ for sentence-level meaning. The coefficients $\gamma_i > 0$ are weighting factors that balance the contributions of each similarity component.

\paragraph{Rule Extraction}

\begin{figure*}[t]
    \centering
    \includegraphics[width=0.8\textwidth]{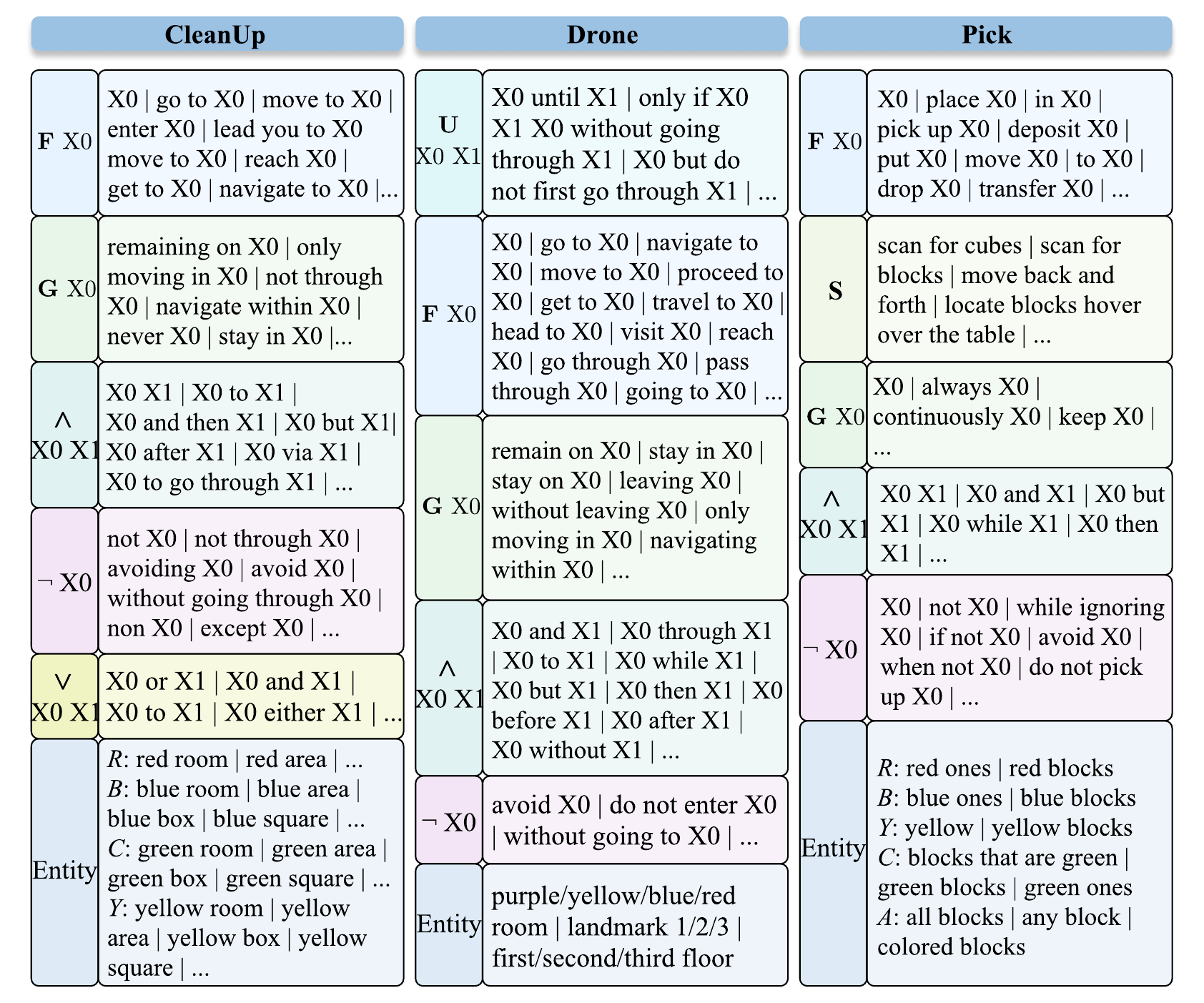}
    \caption{\textbf{Extracted \ac{ltl}-\ac{scfg} rules across domains.} The tables display the mapping between formal \ac{ltl} operators (\eg, $\mathbf{F}, \mathbf{G}, \neg$) and their lexical realizations in different datasets.}
    \label{fig:rules}
\end{figure*}

Upon obtaining the alignment function $\omega_i$, we proceed to extract a set of grammar rules following a Hiero-style hierarchical phrase-based framework~\citep{chiang2005hiero}. This process identifies consistent phrase pairs and recursively composes them into larger hierarchical translation rules. Fig.~\ref{fig:rules} visualizes examples of these extracted grammar rules across three different domains (CleanUp, Drone, and Pick), demonstrating how \ac{nl} phrases map to specific \ac{ltl} operators and \acp{ap}.

Given the assumption that target-side \ac{ltl} expressions contain at most one operator and restrict operands to nonterminal symbols, we can recursively compose rules along the syntax tree of the \ac{ltl} formula. More precisely, a grammar rule is derived from each node $v \in V_{\varphi_i}$ in the syntax tree. Let $s_i(v) \in s_i$ denote the \ac{nl} phrase corresponding to the subformula $\ell_{\varphi_i}(v)$. The extraction follows:
\begin{equation}\label{equ:get_rules}
\left\{
\begin{aligned}
    & \omega_i\left(\lambda_{\varphi_i}\left(v\right)\right) \parallel \lambda_{\varphi_i}(v) \\
    & \quad \Leftrightarrow \lambda_{\varphi_i}(v) \in \mathcal{P}, \\
    & p_v\big(\omega_i(\lambda_{\varphi_i}(v)), X_{v_1}\big) \parallel \lambda_{\varphi_i}(v) \ X_{v_1} \\
    & \quad \Leftrightarrow
    \begin{aligned}[t]
        & (v,v_1) \in E_{\varphi_i}, \lambda_{\varphi_i}(v) \in \OpsU,
    \end{aligned} \\
    & p_v\big(X_{v_1}, \omega_i(\lambda_{\varphi_i}(v)), X_{v_2}\big) \parallel \lambda_{\varphi_i}(v) \ X_{v_1} \ X_{v_2} \\
    & \quad \Leftrightarrow
    \begin{aligned}[t]
        & (v,v_1), (v,v_2) \in E_{\varphi_i}, \lambda_{\varphi_i}(v) \in \OpsB,
    \end{aligned}
\end{aligned}
\right.
\end{equation}
where $p_v(L) \in \text{Perm}(L)$ denotes an arbitrary permutation of the corresponding symbols that preserves the lexical order in the \ac{nl} text. The sets $\OpsU$ and $\OpsB$ represent the unary and binary operators supported by the system, respectively. We further assume the following composition property for the source-side phrase:
\begin{equation}\label{equ:assume_all_add}
 \sum_{x \in p_v} [x \cdot \mathbb{I}(x \in \Sigma_\alpha^*) + s_i(u) \cdot \mathbb{I}(x = X_u)] = s_i(v).
\end{equation}

Under this assumption, the rule extraction process can be further optimized. The objective function $J$ can be expanded as:
\begin{equation}\label{eq:J_expand}
\begin{aligned}
J(\omega_i,s_i,T_{\varphi_i})
&= \textstyle\sum_{v}
[
S(\mathrm{MCMC}(v,\omega_i),\, s_B(\ell_{\varphi_i}(v))) \\
&\hspace{3.4em} {}+
S(\omega_i(\lambda_{\varphi_i}(v)),\, f_B(\lambda_{\varphi_i}(v)))
] \\
&= \textstyle\sum_{v}
[
S(s_i(v),\, s_B(\ell_{\varphi_i}(v))) \\
&\hspace{3.4em} {}+
S(\omega_i(\lambda_{\varphi_i}(v)),\, f_B(\lambda_{\varphi_i}(v)))
] \\
&= \textstyle\sum_{v}
S(\omega_i(\lambda_{\varphi_i}(v)),\, f_B(\lambda_{\varphi_i}(v))) \\
&\quad {}+
\textstyle\sum_{v_s \in \text{Son}(v)}
S(s_i(v_s),\, s_B(\ell_{\varphi_i}(v_s))).
\end{aligned}
\end{equation}
This formulation reveals that word alignment and rule extraction can be optimized jointly by refining $s_i(v)$ along the \ac{ltl} syntax tree $T_{\varphi_i}$ using a top-down greedy algorithm. This approach is predicated on the assumption that $s_i(v)$ is composed of its children's phrases $s_i(v_s)$ and the aligned operator text $\omega_i(\lambda_{\varphi_i}(v))$ under a specific permutation. 

By leveraging the simplified optimization described above, we can obtain the set of grammar rules corresponding to Eq. \eqref{equ:get_rules}. Subsequently, the proposed \ac{ltl}-\ac{scfg} can be derived through the merging grammar rules with same target-side operator $o\in \Sigma_\beta$. The complete \ac{ltl}-\ac{scfg} rule extraction procedure is detailed in Alg.~\ref{alg:extraction}.

\subsection{Hierarchical Decision-Making Translation}

This section presents how the \ac{ltl}-\ac{scfg} constraints are integrated into the \ac{rl} exploration process within our translation model. We first formulate \ac{ltl} translation as an \ac{mdp}. Our framework constrains the agent's decision space according to \ac{ltl}-\ac{scfg} rules by decoupling high-level decision-making from subtask decomposition. Subsequently, we discuss the optimization process, where the agent is trained through \ac{rl} with teacher forcing, which improves exploration efficiency while preserving translation quality.

\paragraph{Translation as a Markov Decision Process}\label{subsubsubsec:mdp}

The process of translating \ac{ltl} can be modeled as a sequential decision-making task: selecting the outermost corresponding operator or \acp{ap} in the \ac{nl} $\mathbf{o} \in\Sigma_\beta$ and applying the corresponding \ac{ltl}-\ac{scfg} rule to transition into different branch states (the sub-sentences $\{\hat{s}({v}_s)\}$ corresponding to $\{{v}_s\}$ in the translation model). Hence, the translation process can be expressed in a top-down, tree-structured manner as
\begin{equation}
    \mathbf{o}, \{\hat{s}({v}_s)\} = \pi(v, \hat{s}(v), T_\varphi).
\end{equation}
Accordingly, we define that the translation model receives a fixed reward for correct primitive $r_v = \epsilon \times[\mathbf{o} = \lambda_\varphi(v)]$. The optimization objective of the translation model can thus be expressed as
\begin{subequations}\label{equ:opt_trans}
\begin{align}
\pi^\star
&= \arg\max_{\pi}\,
\mathbb{E}_{(s_i,\varphi_i)\sim\mathcal{D},\,\pi}
\big[\,U\big(s_i(\mathbf{r}_{\varphi_i})\big)\big],
\label{equ:opt_trans_obj}\\
U(\hat{s}(v))
&= r_v
 + \frac{\gamma}{\lvert\mathcal{N}(v)\rvert}
   \sum_{v_s\in\mathcal{N}(v)} U\big(\hat{s}(v_s)\big),
\label{equ:opt_trans_U}
\end{align}
\end{subequations}
where $\gamma$ denotes the discount factor.

We define an \ac{mdp} $\mathcal{M} = \langle \mathcal{S}, \mathcal{A}, P, R, \gamma \rangle$, where $\mathcal{S} = \{s_t = (v, \hat{s}) \mid v\in T_\varphi, \hat{s}\in \Sigma_\alpha^*\}$ is the set of states, and $\mathcal{A} = \{a_t = (\mathbf{o}, \{\hat{s}(v_s)\}) \mid \mathbf{o} \in \Sigma_\beta, \hat{s}(v_s)\in \Sigma_\alpha^*, (v, v_s)\in E_\varphi\}$ is the set of actions. And
\begin{equation}\label{equ:prob}
\begin{aligned}
P\Bigl(s_{t+1}=(v_s,\hat{s}(v_s))\ \Big|\ 
&s_t=(v,\hat{s}(v)),\\
& a_t=\bigl(\mathbf{o},\{\hat{s}(v_s)\}\bigr)\Bigr)
= \frac{1}{\lvert\{v_s\}\rvert},
\end{aligned}
\end{equation}
where $P(s_{t+1} \mid s_{t}, a_t)$ denotes the transition probability from state $s_{t}$ to $s_{t+1}$ under action $a_t$, $R(s_t,a_t) = r_t = \epsilon \times[\mathbf{o}_t = \lambda_{t}(v_t)]$ represents the expected immediate reward for taking action $a_t$ in state $s_t$, and $\gamma \in [0,1]$ is the discount factor that balances immediate and future rewards.

The goal of the translation agent interacting with this \ac{mdp} is to find a policy $\pi(a_t\mid s_t)$ that maximizes the expected discounted return:
\begin{equation}\label{equ:opt_mdp}
\arg\max_\pi \ \mathbb{E}_\pi \left[ \sum_{t=0}^{\infty} \gamma^t r_{t} \right].
\end{equation}
It can be proved that the objective of the \ac{mdp} optimization strategy in Eq. \eqref{equ:opt_mdp} is equivalent to optimizing the translation model in Eq. \eqref{equ:opt_trans}, \ie, the translation task can be reformulated as an equivalent MDP.

\begin{proof}
Let $U_t = \sum_{k=0}^{\infty} \gamma^k r_{t+k}$ denote the cumulative discounted return at time step $t$. It is easy to see that $U_t = r_{t} + \gamma \sum_{k=0}^{\infty}\gamma^k r_{t+k+1} = r_{t} + \gamma U_{t+1}$. Taking the expectation conditioned on $s_t$ and the policy $\pi$, we obtain
\begin{equation*}
\begin{aligned}
\mathbb{E}_\pi[U_t \mid s_t] & = \mathbb{E}_\pi[r_{t} \mid s_t] + \gamma\,\mathbb{E}_\pi[U_{t+1}\mid s_t] \\
& = r_{v} + \gamma\,\mathbb{E}_\pi[U_{t+1}\mid s_t] \\
& = r_{v} + \frac{\gamma}{|\{v_s\}|}\sum_{v_s}\mathbb{E}_\pi[U_{t+1}\mid s_{t+1}].
\end{aligned}
\end{equation*}
This recursive relationship aligns exactly with the definition of $U(\hat{s}(v))$ in Eq. \eqref{equ:opt_trans}. Therefore, $\mathbb{E}_\pi[U_t \mid s_t] = \mathbb{E}_\pi[U(\hat{s}(v))]$.
\begin{equation*}
\begin{aligned}
\mathbb{E}_\pi\!\left[\sum_{t=0}^{\infty}\gamma^t r_t\right]
&= \mathbb{E}_\pi\!\left[U_0 \mid s_0\right] \\
&= \mathbb{E}_{s_0,\pi}\!\left[U_0\right] \\
&= \mathbb{E}_{(s_i,\varphi_i),\,\pi}\!\left[U\big(s_i(\mathbf{r}_{\varphi_i})\big)\right].
\end{aligned}
\end{equation*}
\end{proof}

\paragraph{Self-Constrained Reasoning Language Agent}

Inspired by human cognitive processes, we design a hierarchical translation framework that effectively integrates the constraints into the translation model:
\begin{align}
    \mathbf{o} &= \pi^1(s_t = (v, \hat{s})), \\
    \{\hat{s}({v}_s)\} &= \pi^2(s_t, \mathbf{o}, f(\mathbf{o})).
\end{align}
This approach decouples decision-making from subtask decomposition, enabling the model to operate within a constrained space and to produce evidence-based outputs.

The model first conducts action selection which generates an output decision under the constraints of the \ac{ltl}-\ac{scfg} grammar, which restricts the number of possible actions $|\Sigma_\beta|$. These constraints narrow the search space for the \ac{rl} agent, ensuring that the exploration remains focused on valid and meaningful actions. In practice, $\pi^1$ serves as the target for \ac{rl} training, employing a frozen \ac{bert} model for text embeddings, while the remaining components are encoded in one-hot format. The overall network follows an MLP architecture, with additional history state $h_t = \{s_{t-1}, s_{t-2}, \dots\}$ information incorporated.

Following the action-selection phase, the model performs split execution which decomposes the overall translation task into smaller subtasks guided by the \ac{ltl}-\ac{scfg} constraints. This submodel acts as a frozen module. We trained a neural segmenter based on \ac{bert}, $P_{\text{Bert}}(\{b_i\} \mid \hat{s}, \mathbf{o})$, where the segmentation boundaries $\{b_i\}$ divide a sentence into three parts. To further improve segmentation accuracy, we measure the structural completeness of candidate segments by evaluating their semantic similarity against canonical back-translations derived from the grammar:
\begin{equation}\label{eq:pscfg}
P_{\mathrm{Scfg}}(\{b_i\} \mid \hat{s}, \mathbf{o}) = \frac{F(\{b_i\} \mid \hat{s}, \mathbf{o})}{\sum_{\{b_k\}} F(\{b_k\} \mid \hat{s}, \mathbf{o})},
\end{equation}
where the scoring function $F(\{b_i\} \mid \hat{s}, \mathbf{o})$ evaluates the optimal syntactic alignment between the segmented text spans and the right-hand side components of an \ac{scfg} rule:
\begin{equation}\label{eq:F_def}
F(\{b_i\} \mid \hat{s}, \mathbf{o}) = \max_{m, p\in \text{P}} \Big( \log w_m \times \textstyle\sum_{j} S\big(\hat{s}_{p(j)}, s_B(x_j^{(m)})\big) \Big).
\end{equation}
Here, $w_m\geq1$ denotes the prior weight of rule $\alpha_m \in f(\mathbf{o})$, $p \in \text{P}$ represents a permutation that matches the segmented textual spans $\{\hat{s}_j\}$ to the components $\{x_j^m\} \in \alpha_m$, and $s_B(x_j)$ represents the canonical structured English sentence for sub-clause $x_j^m$. The similarity function $S(\cdot, \cdot)$ utilizes the metrics defined in Eq. \eqref{equ:similarity} to measure semantic alignment.

To balance domain-specific performance and generalization capability, the weights assigned to the neural network and the grammar-based rules are adaptively combined, $P_{\text{Seg}} = \eta_1 P_{\text{Bert}} + \eta_2 P_{\text{Scfg}}$. The subtask decomposition model can be defined as
\begin{equation}\label{eq:seg_argmax}
\begin{aligned}
\{b_i\} &= \arg\max_{\{b_i\}} P_{\mathrm{Seg}}, \\
\hat{s}(v_s) &= \arg\max_{\hat{s}_j \in \hat{s}[b_i, b_{i+1}]} S\big(s_j, s_B(v_s)\big).
\end{aligned}
\end{equation}

\begin{algorithm}[t]
\caption{Extraction of \ac{ltl}-\ac{scfg}}
\label{alg:extraction}
\KwIn{Corpus $C = \{(s_i, \varphi_i)\}$, Back translation rules $f_B$, Optimization function parameters $\gamma_i$}
\KwOut{Extracted \ac{ltl}-\ac{scfg} rules $f$}
\SetKwFunction{FExtractRules}{Extract\_Rules}
\SetKwProg{Fn}{Function}{:}{}

\Fn{\FExtractRules{$s, v, T$}}{
    $\beta \gets \lambda(v)$\;
    $\omega(\lambda(v)), \{s(v_s)\} \gets \arg\max S\left(\omega(\lambda(v)), f_B(\lambda(v))\right) + \sum_{v_s} S\left(s(v_s), s_B(\ell_{\varphi}(v_s))\right)$ \;
    \textbf{s.t.} $\exists p \in \text{Perm}(\omega(\lambda(v)), \{s(v_s)\}), \sum_{x \in p} x = s$\;
    Get $\alpha$ according to Eq. \eqref{equ:get_rules} and Eq. \eqref{equ:assume_all_add}\;
    $R \gets \{(\alpha, \beta)\}$\;
    \ForEach{$(v, v_s) \in E$}{
        $R \gets R \cup \text{\FExtractRules}(s(v_s), v_s, T)$\;
    }
    \Return $R$\;
}

$f \gets \emptyset$\;
\ForEach{$(s_i, \varphi_i) \in C$}{
    Build syntax tree $T_{\varphi_i}$ according to Eq. \eqref{equ:build_tree}\;
    \ForEach{$(\alpha, \beta) \in \text{\FExtractRules}(s_i, \mathbf{r}_{\varphi_i}, T_{\varphi_i})$}{
        $f(\beta)[\alpha] \gets f(\beta)[\alpha] + 1$\;
    }
}
\Return $f$\;
\end{algorithm}

\paragraph{Reinforcement Learning Training}

To optimize the translation decision agent $\pi^1$ under the \ac{ltl}-\ac{scfg} constraints, we adopt a \ac{rl} framework based on \ac{ppo} \citep{schulman2017proximal}. The objective is to maximize the expected cumulative reward defined by the \ac{mdp} described in Sec. \nameref{subsubsubsec:mdp}. Let $\pi^1_\theta$ denote the agent's policy parameterized by $\theta$; the PPO objective with a clipped surrogate is expressed as:
\begin{equation}
\mathcal{L}^{\text{PPO}}(\theta) = \mathbb{E}_t \left[ \min \left( r_t(\theta) \hat{A}_t, \text{clip}(r_t(\theta), 1-\epsilon, 1+\epsilon) \hat{A}_t \right) \right],
\end{equation}
where $r_t(\theta) = \frac{\pi^1_\theta(a_t \mid s_t)}{\pi^1_{\theta_{\text{old}}}(a_t \mid s_t)}$ is the probability ratio, $\hat{A}_t$ is the advantage estimate at timestep $t$, and $\epsilon$ is a small clipping parameter that ensures stable updates. The advantage function $\hat{A}_t$ is computed using the \ac{gae} formulation:
\begin{equation}\label{eq:gae}
\begin{aligned}
\hat{A}_t &= \delta_t + (\gamma\lambda)\delta_{t+1} + \dots + (\gamma\lambda)^{T-t-1}\delta_{T-1}, \\
\delta_t &= r_t + \gamma V(s_{t+1}) - V(s_t),
\end{aligned}
\end{equation}
where $V(s_t)$ denotes the value function, $\gamma$ is the discount factor, and $\lambda$ controls the bias-variance trade-off in advantage estimation.

\begin{figure*}[t]
    \centering
    \includegraphics[width=0.8\textwidth]{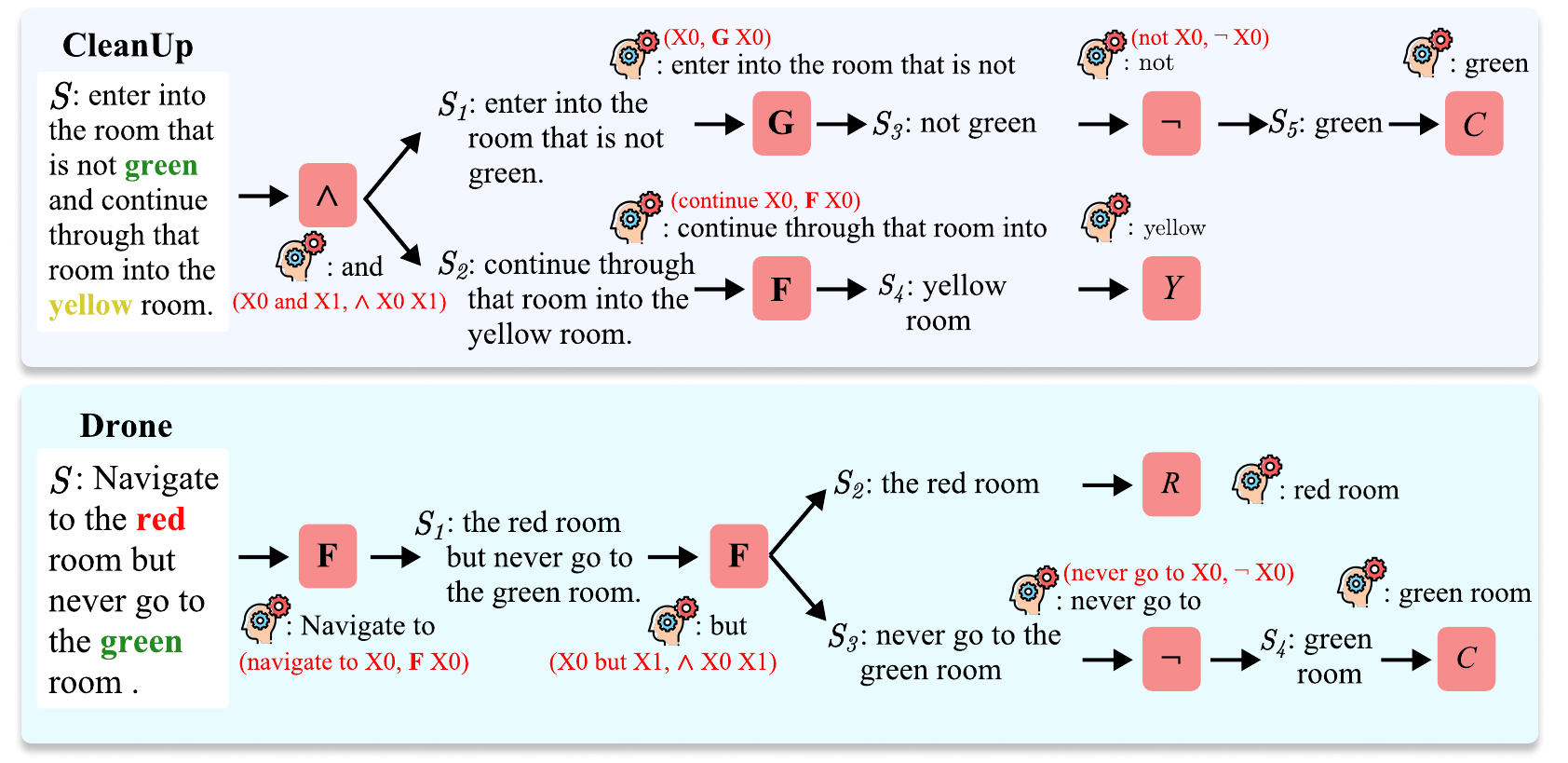}
    \caption{\textbf{Recursive translation showcases.} The diagrams illustrate the step-by-step derivation for CleanUp (top) and Drone (bottom) tasks. The agent recursively selects operators (red blocks) and decomposes the \ac{nl} instruction, yielding a structured and interpretable translation trace.}
    \label{fig:showcase}
\end{figure*}

Given that our translation objective emphasizes the correctness of the current state over long-term rewards in Eq. \eqref{equ:opt_trans}, we incorporate \emph{teacher forcing} during training. Instead of always starting from the root node, the initial state $s_0$ is sampled from any subtree within the \ac{ltl} syntax tree, \ie, $s_0 = (v_s, s(v_s))$. During this process, the agent generates sub-sentences $\hat{s}(v_s)$ under the supervision of the corresponding ground-truth English sentence $s(v_s)$. This strategy facilitates locally consistent decision-making across substructures, ensuring high translation accuracy at each hierarchical level while preserving the global optimization properties of \ac{rl}. Fig.~\ref{fig:showcase} provides step-by-step derivation examples from the CleanUp and Drone domains.

\section{Experiment Setup}

\subsection{Dataset}
To evaluate the capability of our framework to translate \ac{nl} instructions into \ac{ltl} formulas, we utilized three distinct domain-specific datasets \citep{pan2023data}. Each dataset comprises triplets consisting of \ac{nl} instructions, permissible \ac{ap}, and their corresponding target \ac{ltl} formulas. These datasets span diverse application domains, enabling a comprehensive assessment of the framework's performance, its adaptability to heterogeneous contexts, and its ability to generalize mappings from natural language to formal specifications.

Specifically, our evaluation employed three datasets with varying characteristics: Drone Navigation tasks (343 distinct \ac{ltl} formulas), CleanUp tasks (39 distinct \ac{ltl} formulas), and Pick-and-Place tasks (5 distinct \ac{ltl} formulas). For each dataset, the data was partitioned into training and testing sets using a 4:1 ratio. We evaluate the models' capability to satisfy domain constraints on this test set, which is consistently annotated and features the same distribution of \ac{ltl} formulas.

Furthermore, to rigorously assess the framework's ability to generalize to entirely novel scenarios, we established a zero-shot generalization benchmark. This benchmark consists of distinct formulas that were withheld from the training phase, requiring the model to translate instructions without prior exposure to these specific targets. The generalization sets include 51 distinct \ac{ltl} formulas for Drone Navigation and corresponding sentences, 49 for CleanUp, and 3 for Pick-and-Place tasks with 47 different sentences.

\subsection{Baselines}

We categorize the compared \ac{llm} baseline methods into three major classes based on how they incorporate constraints and enhance model capabilities:

\textbf{Knowledge Injection (Soft Constraints):} These methods provide domain-specific knowledge to the model either via in-context learning or parameter updates. This category includes {\ac{llm}-\acs{rag}}, an \ac{llm}-based method incorporating \ac{rag} \citep{lewis2020retrieval} where the optimal shot number is determined via a grid search (range 1--30) on a 10\% validation set, and {fine-tuned models} trained on domain data using the Llama-Factory framework \citep{zheng2024llamafactory} with the \ac{lora} approach \citep{hu2022lora}. Notably, knowledge injection frequently serves as a foundational component that is combined with the subsequent two categories to further augment performance.

\textbf{Decoding Guidance (Hard Constraints):} These methods enforce strict structural or lexical adherence during the generation process. We employ {\ac{gcd}}, implemented via the Guidance framework \citep{willard2023efficient}, which provides a mechanism to inject hard constraints during decoding to ensure output validity.

\textbf{Reasoning and Exploration Enhancement:} These methods aim to improve the model's reasoning and generalization ability. This includes {\ac{cot}} reasoning following the methodology of \citet{wei2022chain}, allowing the model to generate intermediate steps, and {\ac{rl}-based models} built on the simple\ac{rl}-reason framework \citep{zeng2025simplerlzooinvestigatingtamingzero} using a sparse-reward setting where rewards are granted only for entirely correct translations.

Beyond these \ac{llm}-based approaches, we also compare against \textbf{traditional models}, including RNN~\citep{mikolov2010recurrent}, CopyNet~\citep{gu2016incorporating}, \ac{bert}~\citep{devlin2019bert}, and the \ac{bert}-based E-NL2\ac{ltl} model~\citep{pan2023data}, as well as a \textbf{rule-based reference}, a reimplementation of the classical symbolic Hiero framework~\citep{chiang2005hiero}.

\subsection{Training and Evaluation Settings}

For evaluation on public datasets and generalization tests, we adopted the \ac{em} metric as the primary measure~\citep{rajpurkar2016squad}. This metric serves as the standard benchmark, reflecting the requirement that robot-executable translations must be fully accurate to ensure operational safety. We evaluate the model's constraint satisfaction on the domain constraint test set and its generalization capability on the generalization test set, both using the \ac{em} metric. To prevent false negatives arising from syntactically distinct yet semantically equivalent expressions (\eg, $a \land b$ vs. $b \land a$), we apply commutative normalization to reorder operands canonically.

For the human evaluation experiments, we employed metrics commonly used in studies of interpretable models, including \textit{Comprehensibility}, \textit{Trust}, \textit{Transparency}, \textit{Fidelity}, \textit{Interactivity}, and \textit{Decision support}.
\begin{itemize}
    \item \textbf{Comprehensibility}: I can fully understand the logic of the explanation.
    \item \textbf{Trust}: I completely trust the model because of the explanation.
    \item \textbf{Transparency}: I believe the explanation is sufficient for me to clearly understand the model's internal mechanisms or reasoning process.
    \item \textbf{Fidelity}: I believe the explanation fully aligns with my expectations.
    \item \textbf{Interactivity}: I believe I can fully interact with, influence, or explore the model's behavior based on the explanation.
    \item \textbf{Decision Support}: I can fully trust the robot to carry out the next step of the lower-level plan according to the top-level plan, because I believe the explanation significantly reduces the risk of errors caused by incorrect instruction translation.
\end{itemize}
The final human evaluation score was computed by averaging ratings across these six dimensions. In the behavioral experiment, examples were randomly sampled from three distinct scenarios. The explanation processes generated by the interpretable models were standardized and refined into textual form through an \ac{llm}. Each participant completed a Likert-scale questionnaire assessing their responses to six statements, with 1 representing ``strongly disagree,'' 9 representing ``strongly agree,'' and 5 representing ``neutral.'' A total of 12 participants took part in the experiments.

All model outputs were verified to contain no harmful or sensitive content. The evaluation protocol of this study was approved by the Institutional Review Board (IRB) of Peking University. We adhered to strict ethical standards to safeguard the rights and welfare of all participants.

Twelve participants, none of whom had prior experience with robot planning or \ac{ltl}-related knowledge, took part in the behavioral experiments. Each participant was compensated at a rate of \$22.5 per hour. Informed consent was obtained from all participants after providing a detailed explanation of the study's purpose, procedures, potential risks and benefits, and their right to withdraw at any time without penalty. All personal information, including name, age, gender, institution, and educational background, was anonymized and handled in compliance with relevant privacy laws and regulations.

In addition to baseline comparisons, we conducted extensive ablation studies. We ablated our proposed constraint extraction algorithm and compared the results with both the Hiero-based traditional extraction method and unconstrained models. We further examined the contribution of the \ac{rl} component by comparing our constraint-based model with a purely statistical translation model and a fine-tuned action classifier.

During experiments, GPT-4o was used for sampling the back-translation grammar $G_B$. For the sentence similarity function $S$ in Eq.~\eqref{equ:similarity}, all $\gamma_i$ values were set to 1. To balance long-term rewards and variance, the \ac{mdp} discount factor was set to $\gamma = 1$. In constraint extraction, for public dataset evaluation we emphasized the segmenter output by setting $\eta_1 = 0.75$ and $\eta_2 = 0.25$; for generalization testing, we reversed these weightings ($\eta_1 = 0.25$, $\eta_2 = 0.75$).

Except for the \ac{rl}-based \ac{llm} baseline, all experiments were conducted on a single NVIDIA RTX~3090 GPU. For our framework, the batch size was set to 128, with a total of $3 \times 10^6$ time steps and a constant learning rate of $1 \times 10^{-5}$. The action decision module $\pi^1$ was implemented as an \ac{mlp} fine-tuned on top of the frozen \ac{bert} output concatenated with historical context features, and a similar model was used for the segmentation component $P_{\text{Bert}}$.

\section{Experiment Results}

\subsection{Domain Constraint Satisfaction Evaluation}

\begin{table}[t]
\centering
\renewcommand{\arraystretch}{1} 
\caption{\textbf{Domain constraint satisfaction (\%) comparison.} Distinct background colors indicate different method categories.}
\label{tab:main-comparison}
\resizebox{\columnwidth}{!}{
\begin{tabular}{l l S[table-format=2.2] S[table-format=2.2] S[table-format=2.2]}
\toprule
\textbf{Model} & \textbf{Method} & {\textbf{Drone}} & {\textbf{CleanUp}} & {\textbf{Pick}} \\
\midrule
\rowcolor{cllm} Gemini-3 & \ac{rag} & 87.31 & 89.79 & 73.65 \\
\rowcolor{cExpl} Gemini-3 & \ac{rag}+\ac{cot} & 80.76 & 85.36 & 73.65 \\
\rowcolor{cExpl} Gemini-3-think & \ac{rag}+\ac{rl} & 79.55 & 90.38 & 71.62 \\
\rowcolor{cllm} Deepseek-V3 & \ac{rag} & 79.71 & 91.72 & 82.43 \\
\rowcolor{cExpl} Deepseek-V3 & \ac{rag}+\ac{cot} & 74.45 & 86.54 & 85.81 \\
\rowcolor{cExpl} Deepseek-R1 & \ac{rag}+\ac{rl} & 62.73 & 90.98 & 82.43 \\
\rowcolor{cllm} GPT-5 & \ac{rag} & 88.44 & 93.93 & 87.84 \\
\rowcolor{cExpl} GPT-5 & \ac{rag}+\ac{cot} & 73.65 & 93.93 & 85.81 \\
\rowcolor{cExpl} GPT-5-thinking & \ac{rag}+\ac{rl} & 62.65 & 85.50 & 66.22 \\
\rowcolor{cllm} Qwen-3-7B & \ac{rag} & 76.80 & 82.25 & 60.14 \\
\rowcolor{cExpl} Qwen-3-7B & \ac{rag}+\ac{cot} & 51.82 & 65.24 & 47.30 \\
\rowcolor{cCons} Qwen-3-7B & \ac{rag}+\ac{gcd} & 79.79 & 89.79 & 70.95 \\
\rowcolor{cllm} Qwen-3-7B & Finetune & 88.36 & 97.34 & 91.22 \\
\rowcolor{cExpl} Qwen-3-7B & Finetune+\ac{rl} & 87.63 & 90.38 & 73.65 \\
\rowcolor{cCons} Qwen-3-7B & Finetune+\ac{gcd} & 91.59 & 97.49 & 93.92 \\
\rowcolor{cllm} Llama-3-8B & \ac{rag} & 35.33 & 58.28 & 34.46 \\
\rowcolor{cExpl} Llama-3-8B & \ac{rag}+\ac{cot} & 17.30 & 35.95 & 35.14 \\
\rowcolor{cCons} Llama-3-8B & \ac{rag}+\ac{gcd} & 48.10 & 86.24 & 70.95 \\
\rowcolor{cllm} Llama-3-8B & Finetune & 91.59 & 98.22 & 93.92 \\
\rowcolor{cExpl} Llama-3-8B & Finetune+\ac{rl} & 86.90 & 88.91 & 69.59 \\
\rowcolor{cCons} Llama-3-8B & Finetune+\ac{gcd} & 92.24 & 96.75 & 94.59 \\

\addlinespace[2pt]

\rowcolor{cBase} RNN & From Scratch & 87.15 & 95.41 & 93.24 \\
\rowcolor{cBase} CopyNet & From Scratch & 88.92 & 95.41 & 92.57 \\
\rowcolor{cBase} \ac{bert} & Finetune & 91.35 & 97.63 & 92.57 \\
\rowcolor{cBase} E-\acs{nl}2\ac{ltl} & Finetune+\ac{gcd} & 90.70 & 97.78 & 95.95 \\
\rowcolor{cBase} Hiero & Statistical & 30.64 & 30.47 & 27.70 \\

\addlinespace[2pt]

\rowcolor{cOurs} \textbf{Ours} & \ac{scr} & \textbf{94.66} & \textbf{98.82} & \textbf{97.30} \\
\bottomrule
\end{tabular}
}
\end{table}

We evaluated our framework against state-of-the-art methods across three domains: Drone Navigation, CleanUp, and Pick-and-Place. Domain constraint satisfaction was measured using the \ac{em} metric in domain constraint test set~\citep{rajpurkar2016squad}. Tab.~\ref{tab:main-comparison} summarizes the comparison with baseline methods.

\textbf{Overall performance.} Our proposed framework (Ours, \ac{scr}) achieves the best performance across all benchmarks, with constraint satisfaction rates of 94.66\%, 98.82\%, and 97.30\% respectively. It consistently outperforms the strongest baselines, including fine-tuned Llama-3-8B and specialized encoder-based models like E-\acs{nl}2\ac{ltl}, demonstrating superior domain constraint satisfaction in translating \ac{nl} to \ac{ltl}.

\textbf{\ac{llm} with \ac{rag}.} In the few-shot context, \acp{llm} equipped with \ac{rag} show moderate success, with GPT-5 leading this category. However, a critical observation is that \ac{rem} techniques consistently degrade domain constraint satisfaction. For instance, GPT-5's score in the Drone domain drops from 88.44\% to 73.65\% when \ac{cot} is introduced. This suggests that while reasoning chains help with general logic, they often introduce violation of strict domain-specific grammar. In contrast, incorporating \ac{gcd} significantly enhances performance; for Llama-3-8B, \ac{gcd} improves the Pick domain score from a low 34.46\% to 70.95\%, effectively grounding the model's output within valid symbolic boundaries.

\textbf{Fine-tuned Models and \ac{rem} vs. \ac{gcd}.} While domain-specific fine-tuning provides a high baseline, the choice of optimization strategy is crucial. We find that \ac{rem} via unconstrained \ac{rl} (Finetune+\ac{rl}) tends to decrease performance compared to standard fine-tuning (\eg, Llama-3-8B drops from 91.59\% to 86.90\% in the Drone domain). This indicates that without structural priors, \ac{rl} agents may sacrifice grammatical correctness for partial reward signals. Conversely, the application of \ac{gcd} to fine-tuned models provides a consistent performance boost, achieving the highest baseline results (92.24\% for Llama-3-8B in Drone).

\textbf{Specialized and statistical models.} Encoder-based architectures like \ac{bert} and E-\acs{nl}2\ac{ltl} (which utilizes \ac{gcd}) maintain robust performance, particularly in the CleanUp domain (97.78\%). However, purely statistical methods (Hiero) fail to scale, with satisfaction rates hovering around 30\%. This underscores the necessity of combining high-capacity neural models with explicit structural constraints to solve complex formal translation tasks.

\begin{table}[t]
\centering
\caption{\textbf{Ablation study results.}}
\label{tab:ablation}
\resizebox{0.8\columnwidth}{!}{
\begin{tabular}{lcccc}
\hline
\textbf{Model} & \textbf{Drone} & \textbf{CleanUp} & \textbf{Pick} \\
\hline
Ours w/o \ac{ltl}-\ac{scfg} & 87.15 & 97.34 & 95.95 \\
Ours w/o Hierarchical & 89.73 & 92.90 & 89.19 \\
\textbf{Ours} & \textbf{94.66} & \textbf{98.82} & \textbf{97.30} \\
\hline
\end{tabular}%
}
\end{table}

\begin{table}[t]
\centering
\caption{\textbf{Generalization performance (\%) comparison.}}
\label{tab:generalization}
\resizebox{\columnwidth}{!}{
\begin{tabular}{lcccc}
\hline
\textbf{Model Architecture} & \textbf{Method} & \textbf{Drone} & \textbf{Cleanup} & \textbf{Pick} \\
\hline
\rowcolor{cllm} GPT-5 & \ac{rag} & 45.10 & 26.53 & 38.30 \\
\rowcolor{cExpl} GPT-5 & \ac{rag}+\ac{cot} & 43.14 & 32.65 & 68.09 \\
\rowcolor{cExpl} GPT-5-thinking & \ac{rag}+\ac{rl} & 45.10 & 40.82 & 68.09 \\
\rowcolor{cllm} Qwen-3-7B & \ac{rag} & 23.53 & 14.29 & 31.91 \\
\rowcolor{cExpl} Qwen-3-7B & \ac{rag}+\ac{cot} & 29.41 & 20.41 & 57.45 \\
\rowcolor{cCons} Qwen-3-7B & \ac{rag}+\ac{gcd} & 19.61 & 4.08 & 14.89 \\
\rowcolor{cllm} Qwen-3-7B & Finetune & 41.18 & 10.20 & 59.57 \\
\rowcolor{cExpl} Qwen-3-7B & Finetune+\ac{rl} & 41.18 & 34.69 & 68.09 \\
\rowcolor{cCons} Qwen-3-7B & Finetune+\ac{gcd} & 31.37 & 8.16 & 59.57 \\
\rowcolor{cBase} E-\acs{nl}2\ac{ltl} & Finetune+\ac{gcd} & 33.33 & 2.04 & 36.17 \\
\rowcolor{cBase} Hiero & Statistic & 5.88 & 10.20 & 10.64 \\
\rowcolor{cOurs} \textbf{Ours} & SC+\ac{rl} & \textbf{54.90} & \textbf{48.98} & \textbf{74.47} \\
\hline
\end{tabular}%
}
\end{table}

\begin{table*}[t]
\centering
\caption{\textbf{Safety violation comparison across different models and tasks} using \acf{pf}, \acf{ua}, and overall \acf{sv} metrics (\%).}
\label{tab:model_comparison}
\resizebox{0.8\textwidth}{!}{
\begin{tabular}{ll ccc ccc ccc}
\toprule
\multirow{2}{*}{\textbf{Model}} & \multirow{2}{*}{\textbf{Method}} & \multicolumn{3}{c}{\textbf{Drone}} & \multicolumn{3}{c}{\textbf{Cleanup}} & \multicolumn{3}{c}{\textbf{Pick}} \\
\cmidrule(lr){3-5} \cmidrule(lr){6-8} \cmidrule(lr){9-11}
& & \textbf{PF} & \textbf{UA} & \textbf{SV} & \textbf{PF} & \textbf{UA} & \textbf{SV} & \textbf{PF} & \textbf{UA} & \textbf{SV} \\
\midrule
\rowcolor{cllm} GPT-5 & \ac{rag} & 7.19 & 8.57 & 10.75 & 1.92 & 1.78 & 3.11 & 2.03 & 8.78 & 9.46 \\
\rowcolor{cExpl} GPT-5 & \ac{rag}+\ac{cot} & 22.23 & 22.64 & 25.46 & 2.22 & 1.48 & 2.96 & 0.00 & 13.51 & 13.51 \\
\rowcolor{cExpl} GPT-5-thinking & \ac{rag}+\ac{rl} & 31.93 & 29.91 & 35.00 & 10.50 & 7.84 & 11.39 & 21.62 & 23.65 & 33.11 \\
\rowcolor{cllm} Qwen-3-7B & \ac{rag} & 7.44 & 8.97 & 11.72 & 10.65 & 9.62 & 12.57 & 22.30 & 36.49 & 39.19 \\
\rowcolor{cExpl} Qwen-3-7B & \ac{rag}+\ac{cot} & 31.93 & 31.04 & 35.49 & 31.51 & 14.05 & 32.99 & 36.49 & 45.95 & 48.65 \\
\rowcolor{cCons} Qwen-3-7B & \ac{rag}+\ac{gcd} & 0.00 & 4.04 & 4.04 & 0.00 & 2.51 & 2.51 & 0.00 & 27.03 & 27.03 \\
\rowcolor{cllm} Qwen-3-7B & Finetune & 0.00 & 6.71 & 6.71 & 0.00 & 0.44 & 0.44 & 0.00 & 5.41 & 5.41 \\
\rowcolor{cExpl} Qwen-3-7B & Finetune+\ac{rl} & 6.79 & 9.78 & 11.16 & 5.62 & 7.10 & 7.69 & 4.73 & 27.03 & 27.70 \\
\rowcolor{cCons} Qwen-3-7B & Finetune+\ac{gcd} & 0.00 & 7.60 & 7.60 & 0.00 & 0.59 & 0.59 & 0.00 & 6.08 & 6.08 \\
\rowcolor{cBase} E-NL2\ac{ltl} & Finetune+\ac{gcd} & 0.00 & 7.92 & 7.92 & 0.00 & 0.59 & 0.59 & 0.00 & 6.76 & 6.76 \\
\rowcolor{cBase} Hiero & Statistic & 0.00 & 2.75 & 2.75 & 0.00 & 4.73 & 4.73 & 0.00 & 3.38 & 3.38 \\
\rowcolor{cOurs} \textbf{Ours} & \ac{scr} & \textbf{0.00} & \textbf{1.86} & \textbf{1.86} & \textbf{0.00} & \textbf{0.30} & \textbf{0.30} & \textbf{0.00} & \textbf{2.03} & \textbf{2.03} \\
\bottomrule
\end{tabular}
}
\end{table*}

\subsection{Ablation Study}

We conducted an ablation study (Tab.~\ref{tab:ablation}) to isolate the contributions.

\textbf{Impact of \ac{ltl}-\ac{scfg}.} 
The removal of the \ac{ltl}-\ac{scfg} (Ours w/o \ac{ltl}-\ac{scfg}) resulted in a significant performance drop, particularly in the Drone domain (down from 94.66\% to 87.15\%). Without the internalized constraints to guide the model's action space, the system reverts to a more standard translation process that lacks structural grounding. This confirms that embedding domain knowledge directly into the decision-making dynamics is superior to treating constraints as external, post-hoc filters.

\textbf{Importance of Hierarchical Translation Modeling.} 
The most substantial degradation occurred when we disabled the hierarchical translation mechanism (Ours w/o Hierarchical). In the Pick-and-Place domain, accuracy plummeted by over 8 percentage points (from 97.30\% to 89.19\%). This aligns with our hypothesis that complex \ac{ltl} tasks are inherently multi-stage and require hierarchical reasoning. By forcing a flat, linear token-generation process, the model struggles to maintain long-range temporal dependencies and ordering relations.

\subsection{Generalization Evaluation}

To evaluate robustness against unseen instructions, we further assessed generalization performance on held-out test sets across three domains, each consisting entirely of novel \ac{ltl} formulas. Tab.~\ref{tab:generalization} summarizes the comparative results.

\textbf{The Conflict between Constraint and Generalization.} A key finding is that while \ac{gcd} improves domain constraint satisfaction in the main experiments (Tab.~\ref{tab:main-comparison}), it severely hampers generalization on unseen formulas. For example, Qwen-3-7B using \ac{rag}+\ac{gcd} drops to a mere 4.08\% in the Cleanup domain, significantly lower than its unconstrained \ac{rag} counterpart (14.29\%). This confirms that purely passive constraint enforcement filters out valid reasoning paths when the model encounters novel linguistic structures, stifling its ability to reason beyond the training distribution.

\textbf{\ac{rem} and Generalization.} In contrast, \ac{rem} techniques (\ac{cot} and \ac{rl}) demonstrate clear benefits for generalization. Both GPT-5 and Qwen-3-7B see substantial gains in the Pick domain when \ac{cot} or \ac{rl} is applied, with GPT-5's accuracy jumping from 38.30\% to 68.09\%. This suggests that unconstrained reasoning allows \acp{llm} to leverage their internal logic to decompose complex, unseen instructions.

\textbf{Synergy of Self-Constrained Reasoning.} Our framework (Ours) effectively resolves this tension by achieving the highest generalization scores across all domains (54.90\%, 48.98\%, and 74.47\%). Unlike \ac{gcd}, which acts as a rigid filter, our approach internalizes the \ac{ltl}-\ac{scfg} within the decision-making process. This allows the \ac{rl} agent to explore a wide range of reasoning strategies for novel instructions while ensuring that every exploration step remains within the bounds of valid \ac{ltl} syntax.

\subsection{Safety Violation Evaluation}\label{subsec:halluevaluation}

\begin{figure*}[t]
\centering
\includegraphics[width=0.8\textwidth]{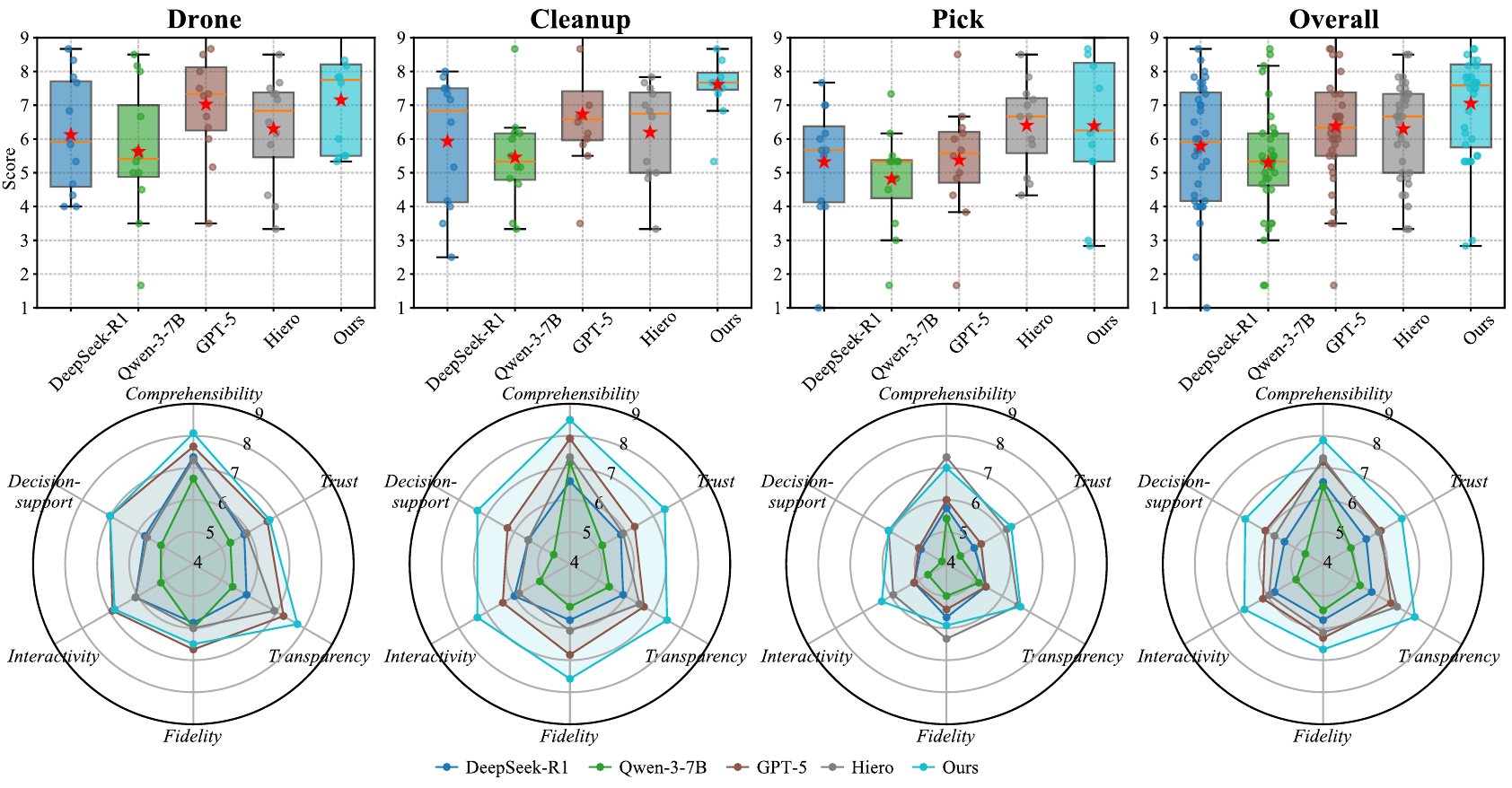}
\caption{\textbf{Human evaluation and multi-dimensional interpretability analysis.} The top row displays boxplots of evaluation scores across the Drone, Cleanup, and Pick domains, along with the Overall performance; red stars indicate mean values. The bottom row presents radar charts comparing our framework against baselines across six qualitative metrics: Comprehensibility, Trust, Transparency, Fidelity, Interactivity, and Decision-support. }
\label{fig:inter}
\end{figure*}

\begin{figure*}[t]
\centering
\includegraphics[width=\textwidth]{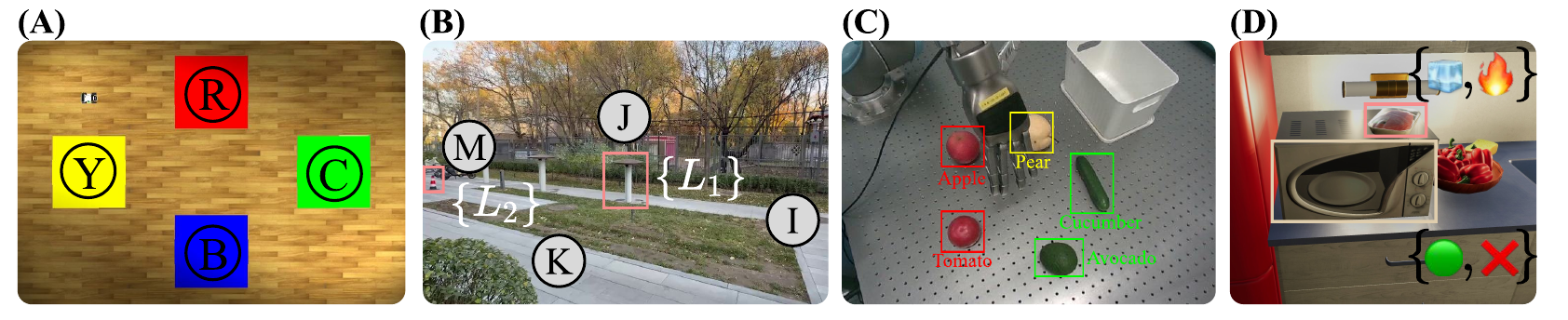}
\caption{\textbf{Illustration of application experiment environments.} \textbf{(A)} The room cleaning environment features a rectangular wood-grain floor divided into four semantic rooms, each labeled by its color. A red room marks the upper-left corner, a yellow room the lower-left, a green room the middle-right, and a blue room the bottom-center. These colored rooms serve as coordinate waypoints for robot navigation. \textbf{(B)} The cargo delivery environment is set in an outdoor pedestrian area bordered by greenery. The layout includes several key coordinates: starting point I on the right, and waypoint J (labeled Landmark 1, $\{L_1\}$) marked by a rectangular stone pedestal in the upper-left. Landmark 2 ($\{L_2\}$) is located at point M on the left lawn's edge, with an additional waypoint K near the bottom. To reach M from I, the agent must pass through either J or K. \textbf{(C)} The manipulation environment features a gray metallic tabletop with a white rectangular storage basket in the upper-right. Scattered across the table are several objects labeled by color: two red fruits (Apple and Tomato), one yellow Pear, and two green fruits (Avocado and Cucumber). \textbf{(D)} The household environment features a kitchen countertop setup with a microwave on the left and a piece of salmon placed on top. The environment involves the following object states: the salmon can be in either a cold or hot state ($\{$\texttwemoji{ice}$,$\texttwemoji{fire}$\}$), while the microwave can be toggled between an active or inactive status ($\{$\texttwemoji{green_circle}$,$\texttwemoji{x}$\}$).}
\label{fig:expenv}
\end{figure*}

To investigate the impact of translation failures on robot safety, we categorize translation-induced errors into two primary failure modes:
(i)~\acf{pf}, which occur when the synthesized formula violates syntactical constraints, rendering it unparseable by downstream formal planners and leading to system deadlock or execution aborts; and
(ii)~\acf{ua}, which occur when the formula is syntactically valid but maps to non-existent or misaligned environmental entities and behaviors, directly leading to hazardous control decisions and safety violations in the physical workspace.

Tab. \ref{tab:model_comparison} reveals that techniques such as \ac{cot} and \ac{rl} often exacerbate safety violations in formal tasks. For instance, the \ac{pf} rate for GPT-5-thinking surges to 31.93 in the Drone domain, suggesting that unconstrained reasoning frequently generates syntactically invalid structures.

Analysis of safety violations across various models further clarifies the sources of domain constraint violations. While \ac{gcd} eliminates grammatical errors, it fails to improve semantic grounding, as evidenced by Qwen-3-7B’s 27.03\% \ac{ua} rate in the Pick domain. In contrast, our \ac{scr} framework effectively resolves both issues. By internalizing the \ac{ltl}-\ac{scfg}, our method achieves zero \acp{pf} while maintaining the lowest \ac{ua} rates across all domains, reaching as low as 0.30\% in Cleanup and 2.03\% in Pick. These results demonstrate that the \ac{scr} framework provides a formally grounded reasoning space that outperforms both unconstrained \acp{llm} and passive filtering methods.

\subsection{Interpretability Evaluation}

To assess the practical utility of our framework in human-robot collaboration, we conducted a human study evaluating the interpretability and trustworthiness of the generated \ac{ltl} specifications. We compared our approach against four representative baselines: DeepSeek-R1, Qwen-3-7B, GPT-5, and the statistical Hiero model. Participants rated the models on a 1--9 Likert scale across six dimensions.

As illustrated in Fig.~\ref{fig:inter}, our method achieves the highest average scores across all evaluated domains: 7.15 in Drone, 7.61 in CleanUp, and 6.39 in Pick. These results consistently outperform the strongest reasoning-based baseline, GPT-5+\ac{cot}, by a margin of 0.12 to 1.01 points, with statistically significant improvements ($p < 0.05$) observed in the CleanUp and Pick scenarios. This performance gain confirms that the integration of hierarchical structures with formal constraints aligns more effectively with human mental models than unconstrained reasoning chains.

The radar charts further highlight our model's superior cognitive profile, particularly regarding comprehensibility and trust. By decomposing complex temporal logic into verifiable hierarchical steps and internalizing domain constraints via the \ac{ltl}-\acs{scfg}, our approach reduces the execution exception compared to end-to-end models such as DeepSeek-R1, as Sec.~\nameref{subsec:halluevaluation}, thereby providing clearer decision support for human operators. While the Hiero baseline achieves comparable performance in the simpler Pick domain (6.40 vs. 6.39) due to the limited number of primitive formulas and structural depth, the advantages of our self-constrained hierarchical method become critical as task complexity increases. In the Drone and CleanUp domains, our framework remains indispensable for maintaining both high transparency and strict formal correctness.

\subsection{Application Experiments}

This section presents experimental results demonstrating the proposed \ac{ltl} translation framework in simulation and real-world environments. The framework was validated across various embodiments, including wheeled robots, humanoid robots, and dexterous hands. Each experiment was repeated at least three times, achieving a 100\% success rate. The software architecture is implemented on Ubuntu 20.04 and ROS Noetic, running on a remote host equipped with an Intel Xeon Gold 5218 CPU (2.30 GHz), 128 GB RAM, and an NVIDIA RTX 3090 GPU.

\subsubsection{Room Cleaning Experiment}

\begin{figure}[t]
\centering
\includegraphics[width=\linewidth]{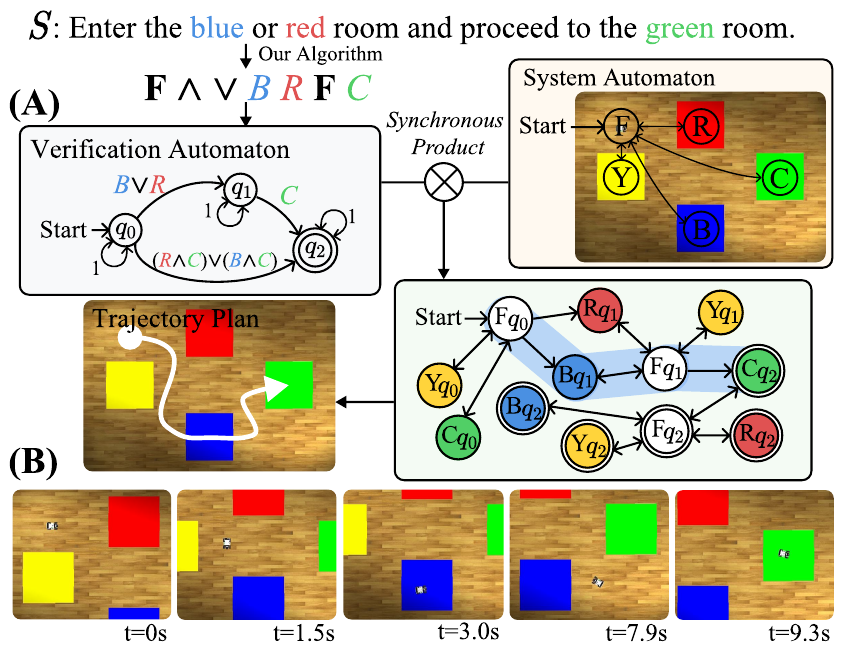}
\caption{\textbf{Translation and execution experiments for the robot room cleaning task.} \textbf{(A)} Upon receiving the natural language instruction $S$, the algorithm translates it into an \ac{ltl} formula. Subsequently, a verification automaton and a system automaton are constructed, and a feasible execution path (indicated by the blue shaded path) is identified by computing their synchronous product. \textbf{(B)} The sequence illustrates the robot's execution trajectory from $t=0$s to $t=9.3$s. Following the trajectory plan, the robot first enters the blue room ($t=3.0$s) and then proceeds to the final objective, the green room ($t=9.3$s), verifying the effectiveness of the algorithm in handling logical instructions and motion planning.}\label{fig:roomclean}
\end{figure}

To further verify the effectiveness of the proposed algorithm in translating natural language instructions with logical relationships and motion planning, we conducted a room cleaning task experiment. As shown in Fig. \ref{fig:expenv}\textbf{(A)}, the simulation environment based on Gazebo consists of four distinct colored rooms, each labeled by its color: the Yellow room (labeled by $Y$), Blue room (labeled by $B$), Red room (labeled by $R$), and Green room (labeled by $C$). 

The system can be defined by a labeled \ac{fts} $\mathcal{T} = (\mathbf{S}, \Sigma, \mathcal{P}, \rightarrow, \mathbf{S}_0, \mathbf{L})$, where $\mathbf{S}=\{\text{F}, \text{Y}, \text{B}, \text{R}, \text{C}\}$ is a finite set of states representing the robot's presence in the four colored rooms and a floor area ($\text{F}$); $\Sigma=\{a_{i,\text{F}}, a_{\text{F},i}| i \in \mathbf{S}\}$ is a set of actions denoting the navigation commands to move between a specific room and the floor; $\mathcal{P}=\{Y, B, R, C\}$ is a set of atomic propositions corresponding to the semantic properties of being inside the Yellow, Blue, Red, or Green rooms; $\rightarrow = \{(i, a_{i,\text{F}}, \text{F}), (\text{F}, a_{\text{F}, i}, i) | i \in \mathbf{S}\} \subseteq \mathbf{S} \times \Sigma \times \mathbf{S}$ is a transition relation that models the topological connectivity, implying the robot must pass through the floor area to transition between different rooms; $\mathbf{S}_0=\{\text{F}\} \subseteq \mathbf{S}$ is the set of initial states, indicating the robot starts its task from the wood-grain floor; and $\mathbf{L}=\{(\text{Y}, \{Y\}), (\text{B}, \{B\}), (\text{R}, \{R\}), (\text{C}, \{C\})\}: \mathbf{S} \rightarrow 2^\mathcal{P}$ is a labeling function that maps each physical room state to its corresponding semantic color observation while leaving the floor state $\text{F}$ unlabeled.

Consider a task defined by the natural language instruction $S$: ``Enter the blue or red room and proceed to the green room." Using our algorithm, this instruction is translated into an \ac{ltl} formula, represented symbolically as $\mathbf{\varphi_1} = \mathbf{F} \land\lor~B~R~\mathbf{F}~C$. 

We utilize the \ac{ltl}2BA tool \citep{gastin2001fast} to convert the translated \ac{ltl} formula into a verification automaton $\mathcal{A}_{\varphi_1} = (\mathbf{Q}, 2^\mathcal{P}, \mathbf{d}, \mathbf{Q}_0, \mathbf{F})$ for automated verification. A system behavior satisfies the given task if and only if its execution trace is accepted by this automaton. Here, $\mathbf{Q}=\{q_0, q_1, q_2\}$ is the set of automaton states; $2^\mathcal{P}=2^{\{Y, B, R, C\}}$ is the input alphabet consisting of the power set of the atomic propositions; $\mathbf{d}: \mathbf{Q} \times 2^\mathcal{P} \rightarrow 2^{\mathbf{Q}}$ is the transition relation defined as $\mathbf{d} = \{(q_0, R \lor B, q_1), (q_1, C, q_2), (q_0, (R \land C) \lor (B \land C), q_2), (q_0, \text{1}, q_0), (q_1, \text{1}, q_1), (q_2, \text{1}, q_2)\}$, which dictates the state evolution based on environmental observations; and $\mathbf{Q}_0=\{q_0\}$ and $\mathbf{F}=\{q_2\}$ represent the initial and accepting states, respectively.

To find a feasible plan, we construct the synchronous product of the system model and the verification automaton $\mathbf{P} =\mathcal{T} \otimes \mathcal{A}_{\varphi_1}=(\mathbf{Q}_\mathbf{P}, \Sigma, \mathbf{d}_\mathbf{P}, \mathbf{Q}_{\mathbf{P},0}, \mathbf{F}_\mathbf{P})$, where $\mathbf{Q}_\mathbf{P} = \mathbf{S} \times \mathbf{Q}$ is the set of product states, $\mathbf{d}_\mathbf{P} \subseteq \mathbf{Q}_\mathbf{P} \times \Sigma \times \mathbf{Q}_\mathbf{P}$ is the transition relation defined as $((s, q), a, (s', q')) \in \mathbf{d}_\mathbf{P}$ if $(s, a, s') \in \rightarrow$ and $q' \in \mathbf{d}(q, L(s'))$, $\mathbf{Q}_{\mathbf{P},0} = \{(s_0, q_0) \mid q_0 \in Q_0\}$ is the set of initial states, and $\mathbf{F}_\mathbf{P} = \{(s, q_f) \mid s \in S, q_f \in \mathbf{F}\}$ is the set of accepting states. Then, we can identify a system behavior that satisfies both the transition constraints and the task requirements by searching on $\mathbf{P}$, thereby yielding an executable plan, as shown in Fig. \ref{fig:roomclean}\textbf{(A)}.

As illustrated by the sequence frames in Fig. \ref{fig:roomclean}\textbf{(B)}, the execution strictly follows the generated automaton strategy, beginning at $t=0$s with the robot positioned near the yellow region. According to the executable plan, the robot selects the blue room as its intermediate task between $t=1.5$s and $t=3.0$s. By $t=3.0$s, the robot successfully enters the blue region, thereby satisfying the first logical operator $(B \lor R)$ of the \ac{ltl} formula. Following the visit to the blue room, the plan directs the robot toward the final objective---the green room (C) between $t=7.9$s and $t=9.3$s. At $t=9.3$s, the robot arrives at the green region, marking the successful completion of the global task.

\subsubsection{Cargo Delivery Experiment}

\begin{figure}[t]
\centering
\includegraphics[width=\linewidth]{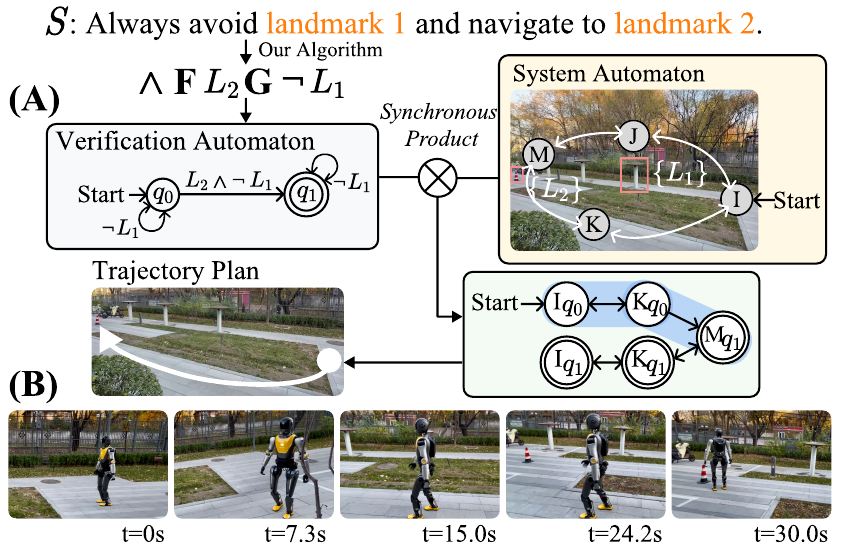}
\caption{\textbf{Translation and execution experiments for the robot cargo delivery task.} \textbf{(A)} The system translates the natural language instruction $S$ into an \ac{ltl} formula. By constructing a verification automaton and a system automaton, a feasible execution plan $\text{I}_{q_0} \rightarrow \text{K}_{q_0} \rightarrow \text{M}_{q_1}$ is identified within the synchronous product (highlighted by the blue shaded path), which strategically bypasses the forbidden waypoint J ($L_1$). \textbf{(B)} The sequence frames illustrate the humanoid robot's real-world trajectory from $t=0$s to $t=30.0$s. To satisfy the safety constraint $\mathbf{G} \neg L_1$, the robot selects the path through waypoint K ($t=15.0$s) and successfully arrives at landmark $L_2$ at point M ($t=30.0$s), demonstrating the algorithm's precision in extracting logical constraints and planning safe motion.}
\label{fig:cargodelivery}
\end{figure}

\begin{figure}[t]
\centering
\includegraphics[width=\linewidth]{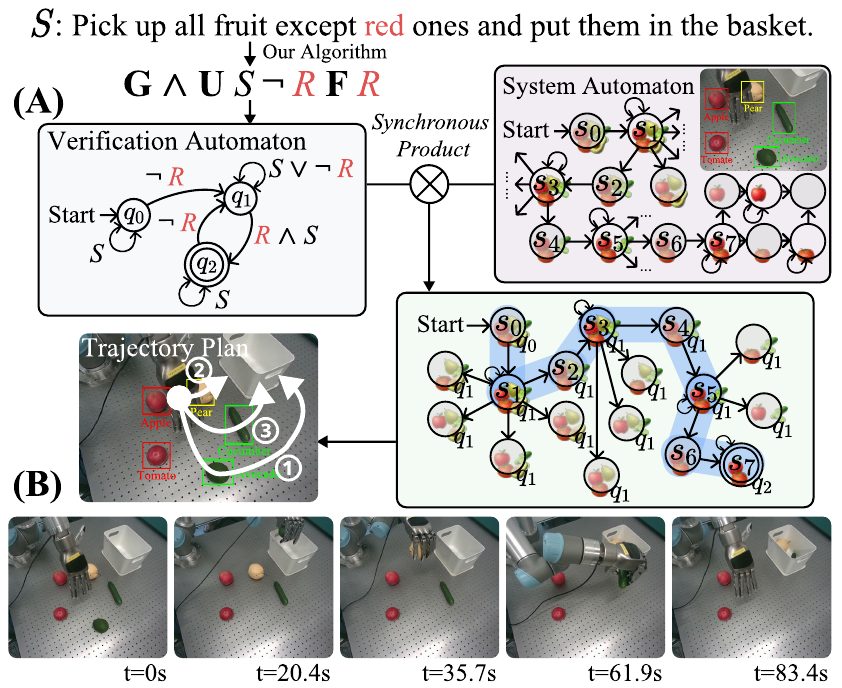}
\caption{\textbf{Translation and execution experiments for the manipulation task.} \textbf{(A)} The system translates the natural language instruction $S$ into the \ac{ltl} formula $\varphi_3$. The synchronous product $\mathbf{P}$ is obtained from the verification automaton $\mathcal{A}_{\varphi_3}$ and the labeled \ac{fts} $\mathcal{T}$. Only the key segments of the system automaton and the synchronous product are illustrated; opaque states within the system automaton denote that those states are unobserved (\ie, without $s$ flag). A feasible execution strategy is identified within the product (highlighted by the blue shaded path), which guides the robot to sequentially interact with non-red objects while satisfying the logical constraints. \textbf{(B)} The trajectory plan and sequence frames illustrate the robotic arm's real-world execution from $t=0$s to $t=83.4$s. The robot selectively picks the avocado ($t=20.4$s), pear ($t=35.7$s), and cucumber ($t=61.9$s), successfully reaching the accepting state $q_2$ and completing the task while leaving the red fruits untouched.}
\label{fig:manipulation}
\end{figure}

To evaluate the capability of the proposed model in extracting logical constraints from natural language instructions and mapping them to formal temporal specifications, we conducted a cargo delivery experiment. As shown in Fig. \ref{fig:expenv}\textbf{(B)}, the experiment is set in an outdoor pedestrian area characterized by a starting point I, a landmark $L_1$ (waypoint J, marked by a stone pedestal), and a destination landmark $L_2$ (point M on the lawn's edge). An additional waypoint K serves as an alternative path.

The environment is modeled as a labeled \ac{fts} $\mathcal{T} = (\mathbf{S}, \Sigma, \mathcal{P}, \rightarrow, \mathbf{S}_0, \mathbf{L})$, where $\mathbf{S}=\{\text{I}, \text{J}, \text{K}, \text{M}\}$ represents the set of waypoints; $\Sigma=\{a_{\text{IK}}, a_{\text{IJ}}, a_{\text{JM}}, a_{\text{KM}}, a_{\text{KI}}, a_{\text{JI}}, a_{\text{MJ}}, a_{\text{MK}}\}$ is the set of navigation actions between adjacent waypoints; $\mathcal{P}=\{L_1, L_2\}$ contains the atomic propositions for the specified landmarks; $\rightarrow=\{(i, a_{ij}, j) | a_{ij} \in \Sigma\}$ defines the topological connectivity, where the agent can move from I to $\{\text{J}, \text{K}\}$ and from $\{\text{J}, \text{K}\}$ to M; $\mathbf{S}_0=\{\text{I}\}$ is the set of initial states; and the labeling function $\mathbf{L}$ maps J to $\{L_1\}$ and M to $\{L_2\}$, with other states being unlabeled.

Consider the natural language instruction $S$: ``Always avoid landmark 1 and navigate to landmark 2." Our algorithm translates this instruction into the \ac{ltl} formula $\varphi_2 = \land~\mathbf{F}~L_2~\mathbf{G}~\neg~L_1$. To verify this task, we construct a verification automaton $\mathcal{A}_{\varphi_2} = (\mathbf{Q}, 2^\mathcal{P}, \mathbf{d}, \mathbf{Q}_0, \mathbf{F})$, where $\mathbf{Q}=\{q_0, q_1\}$ represents the set of internal logic states. The transition relation $\mathbf{d}$ is designed such that the automaton stays in $q_0$ while $\neg L_1$ is satisfied and transitions to the accepting state $q_1$ only when $L_2$ is reached without having ever visited $L_1$.

By computing the synchronous product $\mathbf{P} = \mathcal{T} \otimes \mathcal{A}_{\varphi_2}$, we search for an execution path that satisfies the safety constraint. As depicted in the trajectory plan of Fig. \ref{fig:cargodelivery}\textbf{(A)}, the resulting strategy selects the sequence of product states $\text{I}_{q_0} \rightarrow \text{K}_{q_0} \rightarrow \text{M}_{q_1}$, effectively bypassing the forbidden landmark $L_1$.

The real-world execution by the humanoid robot is illustrated in Fig. \ref{fig:cargodelivery}\textbf{(B)}. At $t=0$s, the robot starts at position I. To satisfy the global safety constraint $\mathbf{G} \neg L_1$, the robot navigates towards waypoint K instead of J. By $t=15.0$s, the robot reaches the vicinity of K. Subsequently, it proceeds toward the final objective M. At $t=30.0$s, the robot arrives at landmark $L_2$, triggering the transition to the accepting state $q_1$ and successfully completing the delivery task while strictly adhering to the avoidance constraint throughout the process.

\subsubsection{Manipulation Experiment}

To evaluate the model's capability in processing natural language instructions for object manipulation, we conducted a series of manipulation experiments. As illustrated in Fig. \ref{fig:expenv}(C), the experimental setup consists of a horizontal workspace containing a white storage bin and three types of colored fruits. Their color mappings are defined as: $\mathbf{C}=\{(\text{Apple}, \allowbreak \{R\}), \allowbreak (\text{Tomato}, \allowbreak \{R\}), \allowbreak (\text{Pear}, \allowbreak \{Y\}), \allowbreak (\text{Cucumber}, \allowbreak \{C\}), \allowbreak (\text{Avocado}, \{C\})\}$ where $R, Y,$ and $C$ represent red, yellow, and green, respectively.

The environment is modeled as a labeled \ac{fts} $\mathcal{T} = (\mathbf{S}, \Sigma, \mathcal{P}, \rightarrow, \mathbf{S}_0, \mathbf{L})$, where the state space $\mathbf{S} = 2^{\text{dom}(\mathbf{C}) \cup \{s\}}$ represents the power set of fruits remaining on the tabletop, augmented with an observation flag $s$ that indicates the completion of a sensing action. The set of actions $\Sigma = \{a_i \mid i \in \text{dom}(\mathbf{C})\} \cup \{\text{S}\}$ includes picking a fruit $i$ (denoted as $a_i$) and the state observation action $\text{S}$, while the set of atomic propositions $\mathcal{P} = \{R, Y, C, \neg R, \neg Y, \neg C, S\}$ characterizes the colors of the fruits, where $\neg R, \neg Y,$ and $\neg C$ represent the existence of fruits that are not red, yellow, or green, respectively, and $S$ denotes the observation status. The transition relation $\rightarrow \subseteq \mathbf{S} \times \Sigma \times \mathbf{S}$ is defined by transitions $(x \cup \{i\}, a_i, x \setminus \{s\})$ for $i \in \text{dom}(\mathbf{C})$ and $s \in x$, modeling the removal of a fruit, and $(x, \text{S}, x \cup \{s\})$ for the observation action which sets the flag. The initial state $\mathbf{S}_0 = \{\text{dom}(\mathbf{C})\}$ represents the tabletop with all fruits present and no prior observation. Finally, the labeling function $\mathbf{L}: \mathbf{S} \rightarrow 2^{\mathcal{P}}$ is defined for any state $x \in \mathbf{S}$ as: $\mathbf{L}(x) = \mathbf{C}(x \setminus \{s\}) \cup \{ \neg c \mid \exists i \in x \setminus \{s\}, c \notin \mathbf{C}(i) \} \cup \{ S \mid s \in x \}$, which maps each state to the colors of the remaining fruits and includes the proposition $S$ whenever the observation flag $s$ is present in the state.

Consider the natural language instruction $S$: ``Pick up all fruit except red ones and put them in the basket." Our algorithm translates this instruction into the \ac{ltl} formula $\varphi_3 = \mathbf{G}\land\mathbf{U}~S~\neg~R~\mathbf{F}~R$. To verify the property $\varphi_3$, a verification automaton is automatically constructed as a 5-tuple $\mathcal{A}_{\varphi_3} = (\mathbf{Q}, 2^\mathcal{P}, \mathbf{d}, \mathbf{Q}_0, \mathbf{F})$. Here, $\mathbf{Q}=\{q_0, q_1, q_2\}$ represents the set of internal logic states. The initial and final state sets are defined as $\mathbf{Q}_0=\{q_0\}$ and $\mathbf{F}=\{q_2\}$, respectively. The state transitions are governed by the relation $\mathbf{d}$, defined as: $\mathbf{d}= \{(q_0, S, q_0), (q_0, \neg R, q_1), (q_1, S \lor \neg R, q_1), (q_1, R \land S, q_2), (q_2, \neg R, q_1), (q_2, S, q_2)\}.$ This automaton captures the evolution of the system logic under the propositional constraints defined in $2^\mathcal{P}$.

By computing the synchronous product $\mathbf{P} = \mathcal{T} \otimes \mathcal{A}_{\varphi_3}$, we search for an execution path that satisfies the instruction. As depicted in the trajectory plan of Fig. \ref{fig:manipulation}(A), the resulting strategy selects a sequence of product states that sequentially navigate the robot through the required workspace configurations. Specifically, the execution begins with the observation action $\text{S}$. The path then proceeds to pick the avocado and the pear, which belong to the sets of green ($C$) and yellow ($Y$) fruits, respectively. Since these actions satisfy the $\neg R$ constraint, the system maintains the logic state $q_1$. Finally, the robot picks the cucumber, completing the removal of all non-red fruits. This sequence ensures that the system satisfies the \ac{ltl} formula $\varphi_3$ by reaching the accepting state $q_2$ in the verification automaton.

The real-world execution by the robotic manipulator is illustrated in Fig. \ref{fig:manipulation}\textbf{(B)}. At $t=0$s, the robot begins in the initial configuration with all five fruits present on the tabletop. Following the trajectory plan derived from the synchronous product, the robot first identifies the target objects and picks up the avocado at $t=20.4$s, which satisfies the $\neg R$ constraint and maintains the logic state $q_1$. Subsequently, it proceeds to pick the pear at $t=35.7$s and the cucumber at $t=61.9$s. By $t=83.4$s, all fruits except the red ones (apple and tomato) have been successfully placed into the storage bin. This sequence of actions ensures that the system transitions through states $s_1 \to s_3 \to s_5$ and finally reaches the accepting state $q_2$ at $s_7$, demonstrating the algorithm's effectiveness in parsing linguistic exclusions and executing precise, multi-step manipulation sequences.

\subsubsection{Household Experiment}

\begin{figure}[t]
\centering
\includegraphics[width=\linewidth]{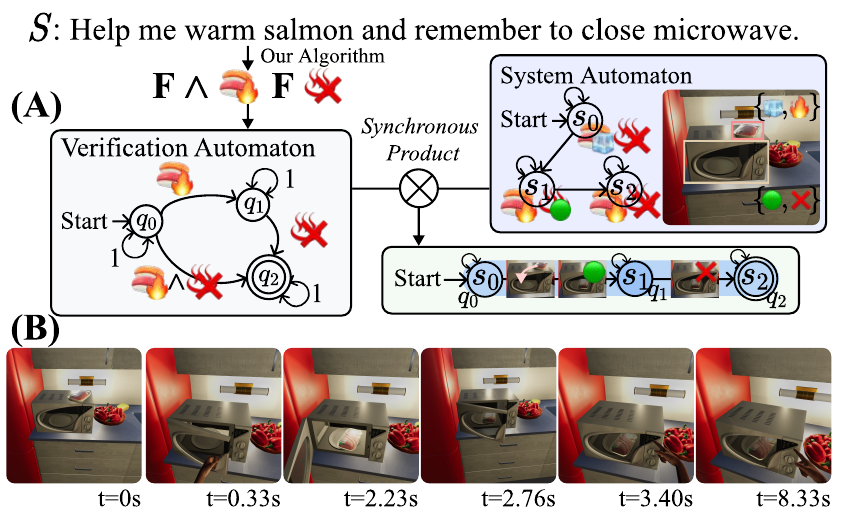}
\caption{\textbf{Translation and execution experiments for the household task.} \textbf{(A)} The system translates the natural language instruction $S$ into an \ac{ltl} formula. By constructing a verification automaton and a system automaton that model the object states, a feasible execution plan $s_{0,q_0} \rightarrow s_{1,q_1} \rightarrow s_{2,q_2}$ is identified within the synchronous product (highlighted by the blue shaded path). This plan guides the agent to successfully heat the food while ensuring the appliance is properly shut off. \textbf{(B)} The sequence frames illustrate the virtual agent's execution trajectory from $t=0$s to $t=8.33$s. Following the generated plan, the agent opens the microwave to insert the salmon ($t=2.23$s), activates it to warm the food ($t=3.40$s), and ultimately deactivates and closes the microwave ($t=8.33$s). This sequence successfully reaches the accepting state $q_2$, demonstrating the algorithm's effectiveness in managing interactions within new action spaces.}
\label{fig:household}
\end{figure}

To evaluate the algorithm's effectiveness in managing interactions within new action spaces, we conducted a household task experiment in the VirtualHome kitchen environment~\citep{puig2018virtualhome}. To adapt our model to this specific domain, we constructed a small dataset comprising 13 natural language sentences and their corresponding \ac{ltl} formulas, covering 12 distinct environmental states. We utilized this dataset to fine-tune our model, which was previously trained on the CleanUp dataset. For the testing phase, we focused on a new specific subset of these states involving the interactions between a piece of salmon and a microwave. 

As depicted in Fig. \ref{fig:expenv}\textbf{(D)}, the household environment features a kitchen countertop. The environment involves the following object states: the salmon's temperature ($\{$\texttwemoji{ice}, \texttwemoji{fire}$\}$) and the microwave's operational status ($\{$\texttwemoji{green_circle}, \texttwemoji{x}$\}$). We model this environment as a labeled \ac{fts} $\mathcal{T} = (\mathbf{S}, \Sigma, \mathcal{P}, \rightarrow, \mathbf{S}_0, \mathbf{L})$. Based on the System Automaton in Fig. \ref{fig:household}, the state space $\mathbf{S}=\{s_0, s_1, s_2\}$ defines their combinations: the initial state $s_0$ (cold salmon, inactive microwave), $s_1$ (hot salmon, active microwave), and $s_2$ (hot salmon, inactive microwave). The action set $\Sigma$ contains physical interaction primitives (\eg, opening, inserting, toggling), while the atomic propositions $\mathcal{P}=\{\texttwemoji{fire}, \texttwemoji{x}\}$ denote the semantic conditions of the food being hot and the appliance being inactive.

Consider the natural language instruction $S$: ``Help me warm salmon and remember to close microwave." Our algorithm translates this instruction into an \ac{ltl} formula $\varphi_4$ that dictates the temporal logic of heating the food and subsequently ensuring the appliance is shut off. To verify this task, a verification automaton $\mathcal{A}_{\varphi_4} = (\mathbf{Q}, 2^\mathcal{P}, \mathbf{d}, \mathbf{Q}_0, \mathbf{F})$ is constructed, where $\mathbf{Q}=\{q_0, q_1, q_2\}$ is the set of logic states. The transition relation dictates that the automaton moves from $q_0$ to $q_1$ when the food is heated (\texttwemoji{fire}), and from $q_1$ to the accepting state $q_2$ when the microwave is closed (\texttwemoji{x}), with an alternative direct transition from $q_0$ to $q_2$ if both conditions are met simultaneously.

By computing the synchronous product $\mathbf{P} = \mathcal{T} \otimes \mathcal{A}_{\varphi_4}$, we search for a feasible execution path. By constructing a verification automaton and a system automaton that model the object states, a feasible execution plan $s_{0,q_0} \rightarrow s_{1,q_1} \rightarrow s_{2,q_2}$ is identified within the synchronous product (highlighted by the blue shaded path in Fig. \ref{fig:household}\textbf{(A)}). This plan sequentially guides the agent to successfully heat the food while ensuring the appliance is properly shut off. 

The simulation execution by the virtual agent is illustrated by the sequence frames in Fig. \ref{fig:household}\textbf{(B)}, tracking the trajectory from $t=0$s to $t=8.33$s. Following the generated plan, the agent starts in the initial state, opens the microwave to insert the salmon at $t=2.23$s, and activates it to warm the food at $t=3.40$s (triggering the transition to $q_1$). Ultimately, the agent deactivates and closes the microwave at $t=8.33$s. By successfully generalizing to this specific task after fine-tuning the model on a small dataset with different states, the execution explicitly demonstrates the algorithm's robust capacity to adapt to and manage entirely new action spaces.

\section{Conclusion}

In this study, we proposed the \ac{scr} framework, which provides an effective way to resolve the inherent tension between reasoning flexibility and formal constraint satisfaction in \ac{nl} to \ac{ltl} translation by internalizing structural knowledge into hierarchical decision-making processes. Unlike traditional post-hoc filtering, our framework integrates an \acs{ltl}-\acs{scfg} constraint representation with reinforcement learning optimization, enabling the model to explore within a formally grounded space to eliminate syntactic errors while significantly enhancing generalization to out-of-distribution instructions. Experimental results demonstrate that the framework consistently outperforms existing baselines in domain-constraint satisfaction, generalization, safety violation reduction, and interpretability, providing a robust and efficient pipeline for transforming human intent into verifiable robotic behavior for complex, long-horizon tasks.

\section*{Acknowledgements}
This work was supported by the National Natural Science Foundation of China (Grant No. 52475001). The authors would like to thank Hecbot Co., Ltd., for providing the humanoid robot platform, and Linkerbot Co., Ltd., for providing the dexterous robotic hand used in this study.

\bibliographystyle{sageh}
\bibliography{references}

@article{agarwal2025think,
  author        = {Agarwal, Bhavik and Joshi, Ishan and Rojkova, Viktoria},
  journal       = {arXiv preprint arXiv:2502.14905},
  title         = {Think inside the json: Reinforcement strategy for strict llm schema adherence},
  year          = {2025}
}

@inproceedings{ahn2022can,
  author        = {ichter, brian and Brohan, Anthony and Chebotar, Yevgen and Finn, Chelsea and Hausman, Karol and Herzog, Alexander and Ho, Daniel and Ibarz, Julian and Irpan, Alex and Jang, Eric and Julian, Ryan and Kalashnikov, Dmitry and Levine, Sergey and Lu, Yao and Parada, Carolina and Rao, Kanishka and Sermanet, Pierre and Toshev, Alexander T and Vanhoucke, Vincent and Xia, Fei and Xiao, Ted and Xu, Peng and Yan, Mengyuan and Brown, Noah and Ahn, Michael and Cortes, Omar and Sievers, Nicolas and Tan, Clayton and Xu, Sichun and Reyes, Diego and Rettinghouse, Jarek and Quiambao, Jornell and Pastor, Peter and Luu, Linda and Lee, Kuang-Huei and Kuang, Yuheng and Jesmonth, Sally and Joshi, Nikhil J. and Jeffrey, Kyle and Ruano, Rosario Jauregui and Hsu, Jasmine and Gopalakrishnan, Keerthana and David, Byron and Zeng, Andy and Fu, Chuyuan Kelly},
  booktitle     = {Proceedings of The 6th Conference on Robot Learning},
  editor        = {Liu, Karen and Kulic, Dana and Ichnowski, Jeff},
  month         = {14--18 Dec},
  pages         = {287--318},
  publisher     = {PMLR},
  series        = {Proceedings of Machine Learning Research},
  title         = {Do As I Can, Not As I Say: Grounding Language in Robotic Affordances},
  volume        = {205},
  year          = {2023}
}

@inproceedings{argenziano2025defining,
  author        = {Argenziano, Francesco and Umili, Elena and Leotta, Francesco and Nardi, Daniele},
  booktitle     = {2025 IEEE 37th International Conference on Tools with Artificial Intelligence (ICTAI)},
  organization  = {IEEE},
  pages         = {1037--1044},
  title         = {Defining and Monitoring Complex Robot Activities via LLMs and Symbolic Reasoning},
  year          = {2025}
}

@book{baier2008principles,
  author        = {Baier, Christel and Katoen, Joost-Pieter},
  publisher     = {MIT press},
  title         = {Principles of model checking},
  year          = {2008}
}

@inproceedings{banerjee2025crane,
  author        = {Banerjee, Debangshu and Suresh, Tarun and Ugare, Shubham and Misailovic, Sasa and Singh, Gagandeep},
  booktitle     = {Proceedings of the 42nd International Conference on Machine Learning},
  editor        = {Singh, Aarti and Fazel, Maryam and Hsu, Daniel and Lacoste-Julien, Simon and Berkenkamp, Felix and Maharaj, Tegan and Wagstaff, Kiri and Zhu, Jerry},
  month         = {13--19 Jul},
  pages         = {2836--2857},
  publisher     = {PMLR},
  series        = {Proceedings of Machine Learning Research},
  title         = {{CRANE}: Reasoning with constrained {LLM} generation},
  volume        = {267},
  year          = {2025}
}

@inproceedings{beurer2024guiding,
  author        = {Beurer-Kellner, Luca and Fischer, Marc and Vechev, Martin},
  booktitle     = {Proceedings of the 41st International Conference on Machine Learning},
  editor        = {Salakhutdinov, Ruslan and Kolter, Zico and Heller, Katherine and Weller, Adrian and Oliver, Nuria and Scarlett, Jonathan and Berkenkamp, Felix},
  month         = {21--27 Jul},
  pages         = {3658--3673},
  publisher     = {PMLR},
  series        = {Proceedings of Machine Learning Research},
  title         = {Guiding {LLM}s The Right Way: Fast, Non-Invasive Constrained Generation},
  volume        = {235},
  year          = {2024}
}

@article{brodo2026property,
  author        = {Brodo, Luca and Scalora, Giuseppe and Henkler, Stefan},
  journal       = {Proceedings of the ACM on Software Engineering},
  number        = {FSE},
  pages         = {2698--2719},
  publisher     = {ACM New York, NY, USA},
  title         = {Property Refinement in Linear Temporal Logic: Formal Semantics and Algorithms for Software Verification},
  volume        = {3},
  year          = {2026}
}

@inproceedings{chen2023nl2tl,
  address       = {Singapore},
  author        = {Chen, Yongchao  and
Gandhi, Rujul  and
Zhang, Yang  and
Fan, Chuchu},
  booktitle     = {Proceedings of the 2023 Conference on Empirical Methods in Natural Language Processing},
  doi           = {10.18653/v1/2023.emnlp-main.985},
  editor        = {Bouamor, Houda  and
Pino, Juan  and
Bali, Kalika},
  month         = {December},
  pages         = {15880--15903},
  publisher     = {Association for Computational Linguistics},
  title         = {{NL}2{TL}: Transforming Natural Languages to Temporal Logics using Large Language Models},
  year          = {2023}
}

@inproceedings{chiang2005hiero,
  address       = {Ann Arbor, Michigan},
  author        = {Chiang, David},
  booktitle     = {Proceedings of the 43rd Annual Meeting of the Association for Computational Linguistics ({ACL}{'}05)},
  doi           = {10.3115/1219840.1219873},
  editor        = {Knight, Kevin  and
Ng, Hwee Tou  and
Oflazer, Kemal},
  month         = {June},
  pages         = {263--270},
  publisher     = {Association for Computational Linguistics},
  title         = {A Hierarchical Phrase-Based Model for Statistical Machine Translation},
  year          = {2005}
}

@book{chomsky2000new,
  author        = {Chomsky, Noam},
  publisher     = {Cambridge University Press},
  title         = {New horizons in the study of language and mind},
  year          = {2000}
}

@book{chomsky2014aspects,
  author        = {Chomsky, Noam},
  publisher     = {MIT press},
  title         = {Aspects of the Theory of Syntax},
  year          = {2014}
}

@article{correa2023humans,
  author        = {Correa, Carlos G and Ho, Mark K and Callaway, Frederick and Daw, Nathaniel D and Griffiths, Thomas L},
  doi           = {10.1371/journal.pcbi.1011087},
  journal       = {PLoS computational biology},
  number        = {6},
  pages         = {e1011087},
  publisher     = {Public Library of Science San Francisco, CA USA},
  title         = {Humans decompose tasks by trading off utility and computational cost},
  volume        = {19},
  year          = {2023}
}

@article{dash2022review,
  author        = {Dash, Tirtharaj and Chitlangia, Sharad and Ahuja, Aditya and Srinivasan, Ashwin},
  journal       = {Scientific Reports},
  number        = {1},
  pages         = {1040},
  publisher     = {Nature Publishing Group UK London},
  title         = {A review of some techniques for inclusion of domain-knowledge into deep neural networks},
  volume        = {12},
  year          = {2022}
}

@inproceedings{devlin2019bert,
  author        = {Devlin, Jacob and Chang, Ming-Wei and Lee, Kenton and Toutanova, Kristina},
  booktitle     = {Proceedings of the 2019 conference of the North American chapter of the association for computational linguistics: human language technologies, volume 1 (long and short papers)},
  pages         = {4171--4186},
  title         = {Bert: Pre-training of deep bidirectional transformers for language understanding},
  year          = {2019}
}

@inproceedings{english2025grammar,
  author        = {English, William H and Simon, Dominic and Jha, Sumit Kumar and Ewetz, Rickard},
  booktitle     = {Proceedings of the 42nd International Conference on Machine Learning},
  editor        = {Singh, Aarti and Fazel, Maryam and Hsu, Daniel and Lacoste-Julien, Simon and Berkenkamp, Felix and Maharaj, Tegan and Wagstaff, Kiri and Zhu, Jerry},
  month         = {13--19 Jul},
  pages         = {15370--15383},
  publisher     = {PMLR},
  series        = {Proceedings of Machine Learning Research},
  title         = {Grammar-Forced Translation of Natural Language to Temporal Logic using {LLM}s},
  volume        = {267},
  year          = {2025}
}

@article{frankland2020concepts,
  author        = {Frankland, Steven M and Greene, Joshua D},
  doi           = {10.1146/annurev-psych-122216-011829},
  journal       = {Annual Review of Psychology},
  pages         = {273--303},
  title         = {Concepts and compositionality: in search of the brain's language of thought},
  volume        = {71},
  year          = {2020}
}

@inproceedings{gastin2001fast,
  author        = {Gastin, Paul and Oddoux, Denis},
  booktitle     = {International Conference on Computer Aided Verification},
  organization  = {Springer},
  pages         = {53--65},
  title         = {Fast LTL to B{\"u}chi automata translation},
  year          = {2001}
}

@article{germiniani2025systematic,
  author        = {Germiniani, Samuele and Nicoletti, Daniele and Pravadelli, Graziano},
  journal       = {IEEE Access},
  publisher     = {IEEE},
  title         = {A Systematic Literature Review on Mining LTL Specifications},
  year          = {2025}
}

@inproceedings{gopalan2018sequence,
  author        = {Gopalan, Nakul and Arumugam, Dilip and Wong, Lawson LS and Tellex, Stefanie},
  booktitle     = {Robotics: Science and Systems},
  pages         = {1--10},
  title         = {Sequence-to-Sequence Language Grounding of Non-Markovian Task Specifications.},
  volume        = {2018},
  year          = {2018}
}

@inproceedings{gu2016incorporating,
  author        = {Gu, Jiatao and Lu, Zhengdong and Li, Hang and Li, Victor OK},
  booktitle     = {Proceedings of the 54th Annual Meeting of the Association for Computational Linguistics (Volume 1: Long Papers)},
  pages         = {1631--1640},
  title         = {Incorporating copying mechanism in sequence-to-sequence learning},
  year          = {2016}
}

@article{gundana2022event,
  author        = {Gundana, David and Kress-Gazit, Hadas},
  journal       = {IEEE Robotics and Automation Letters},
  number        = {4},
  pages         = {10001--10008},
  publisher     = {IEEE},
  title         = {Event-based signal temporal logic tasks: Execution and feedback in complex environments},
  volume        = {7},
  year          = {2022}
}

@inproceedings{guo2025castl,
  author        = {Guo, Weihang and Kingston, Zachary and Kavraki, Lydia E},
  booktitle     = {2025 IEEE International Conference on Robotics and Automation (ICRA)},
  organization  = {IEEE},
  pages         = {11957--11964},
  title         = {Castl: Constraints as specifications through llm translation for long-horizon task and motion planning},
  year          = {2025}
}

@article{guo2025deepseek,
  author        = {Guo, Daya and Yang, Dejian and Zhang, Haowei and Song, Junxiao and Wang, Peiyi and Zhu, Qihao and Xu, Runxin and Zhang, Ruoyu and Ma, Shirong and Bi, Xiao and Zhang, Xiaokang and Yu, Xingkai and Wu, Yu and Wu, Z. F. and Gou, Zhibin and Shao, Zhihong and Li, Zhuoshu and Gao, Ziyi and Liu, Aixin and Xue, Bing and Wang, Bingxuan and Wu, Bochao and Feng, Bei and Lu, Chengda and Zhao, Chenggang and Deng, Chengqi and Ruan, Chong and Dai, Damai and Chen, Deli and Ji, Dongjie and Li, Erhang and Lin, Fangyun and Dai, Fucong and Luo, Fuli and Hao, Guangbo and Chen, Guanting and Li, Guowei and Zhang, H. and Xu, Hanwei and Ding, Honghui and Gao, Huazuo and Qu, Hui and Li, Hui and Guo, Jianzhong and Li, Jiashi and Chen, Jingchang and Yuan, Jingyang and Tu, Jinhao and Qiu, Junjie and Li, Junlong and Cai, J. L. and Ni, Jiaqi and Liang, Jian and Chen, Jin and Dong, Kai and Hu, Kai and You, Kaichao and Gao, Kaige and Guan, Kang and Huang, Kexin and Yu, Kuai and Wang, Lean and Zhang, Lecong and Zhao, Liang and Wang, Litong and Zhang, Liyue and Xu, Lei and Xia, Leyi and Zhang, Mingchuan and Zhang, Minghua and Tang, Minghui and Zhou, Mingxu and Li, Meng and Wang, Miaojun and Li, Mingming and Tian, Ning and Huang, Panpan and Zhang, Peng and Wang, Qiancheng and Chen, Qinyu and Du, Qiushi and Ge, Ruiqi and Zhang, Ruisong and Pan, Ruizhe and Wang, Runji and Chen, R. J. and Jin, R. L. and Chen, Ruyi and Lu, Shanghao and Zhou, Shangyan and Chen, Shanhuang and Ye, Shengfeng and Wang, Shiyu and Yu, Shuiping and Zhou, Shunfeng and Pan, Shuting and Li, S. S. and Zhou, Shuang and Wu, Shaoqing and Yun, Tao and Pei, Tian and Sun, Tianyu and Wang, T. and Zeng, Wangding and Liu, Wen and Liang, Wenfeng and Gao, Wenjun and Yu, Wenqin and Zhang, Wentao and Xiao, W. L. and An, Wei and Liu, Xiaodong and Wang, Xiaohan and Chen, Xiaokang and Nie, Xiaotao and Cheng, Xin and Liu, Xin and Xie, Xin and Liu, Xingchao and Yang, Xinyu and Li, Xinyuan and Su, Xuecheng and Lin, Xuheng and Li, X. Q. and Jin, Xiangyue and Shen, Xiaojin and Chen, Xiaosha and Sun, Xiaowen and Wang, Xiaoxiang and Song, Xinnan and Zhou, Xinyi and Wang, Xianzu and Shan, Xinxia and Li, Y. K. and Wang, Y. Q. and Wei, Y. X. and Zhang, Yang and Xu, Yanhong and Li, Yao and Zhao, Yao and Sun, Yaofeng and Wang, Yaohui and Yu, Yi and Zhang, Yichao and Shi, Yifan and Xiong, Yiliang and He, Ying and Piao, Yishi and Wang, Yisong and Tan, Yixuan and Ma, Yiyang and Liu, Yiyuan and Guo, Yongqiang and Ou, Yuan and Wang, Yuduan and Gong, Yue and Zou, Yuheng and He, Yujia and Xiong, Yunfan and Luo, Yuxiang and You, Yuxiang and Liu, Yuxuan and Zhou, Yuyang and Zhu, Y. X. and Huang, Yanping and Li, Yaohui and Zheng, Yi and Zhu, Yuchen and Ma, Yunxian and Tang, Ying and Zha, Yukun and Yan, Yuting and Ren, Z. Z. and Ren, Zehui and Sha, Zhangli and Fu, Zhe and Xu, Zhean and Xie, Zhenda and Zhang, Zhengyan and Hao, Zhewen and Ma, Zhicheng and Yan, Zhigang and Wu, Zhiyu and Gu, Zihui and Zhu, Zijia and Liu, Zijun and Li, Zilin and Xie, Ziwei and Song, Ziyang and Pan, Zizheng and Huang, Zhen and Xu, Zhipeng and Zhang, Zhongyu and Zhang, Zhen},
  journal       = {Nature},
  number        = {8081},
  pages         = {633--638},
  publisher     = {Nature Publishing Group UK London},
  title         = {DeepSeek-R1 incentivizes reasoning in LLMs through reinforcement learning},
  volume        = {645},
  year          = {2025}
}

@article{guo2026one,
  author        = {Guo, Zijian and I{\c{s}}{\i}k, {\.I}lker and Ahmad, HM and Li, Wenchao},
  journal       = {Advances in Neural Information Processing Systems},
  pages         = {77500--77529},
  title         = {One subgoal at a time: Zero-shot generalization to arbitrary linear temporal logic requirements in multi-task reinforcement learning},
  volume        = {38},
  year          = {2026}
}

@inproceedings{hokamp2017lexically,
  author        = {Hokamp, Chris and Liu, Qun},
  booktitle     = {Proceedings of the 55th Annual Meeting of the Association for Computational Linguistics (Volume 1: Long Papers)},
  pages         = {1535--1546},
  title         = {Lexically constrained decoding for sequence generation using grid beam search},
  year          = {2017}
}

@inproceedings{hu2022lora,
  author        = {Edward J Hu and Yelong Shen and Phillip Wallis and Zeyuan Allen-Zhu and Yuanzhi Li and Shean Wang and Lu Wang and Weizhu Chen},
  booktitle     = {International Conference on Learning Representations},
  pages         = {1--13},
  title         = {Lo{RA}: Low-Rank Adaptation of Large Language Models},
  year          = {2022}
}

@inproceedings{ildizlearning,
  author        = {Muhammed Emrullah Ildiz and Halil Alperen Gozeten and Ege Onur Taga and Samet Oymak},
  booktitle     = {Forty-third International Conference on Machine Learning},
  pages         = {1--31},
  title         = {Learning to Correct: Reinforcement Learning for Multi-Attempt Chain-of-Thought},
  year          = {2026}
}

@article{intelligence2025pi_,
  author        = {Black, Kevin and Brown, Noah and Darpinian, James and Dhabalia, Karan and Driess, Danny and Esmail, Adnan and Equi, Michael and Finn, Chelsea and Fusai, Niccolo and Galliker, Manuel Y. and Ghosh, Dibya and Groom, Lachy and Hausman, Karol and Ichter, Brian and Jakubczak, Szymon and Jones, Tim and Ke, Liyiming and LeBlanc, Devin and Levine, Sergey and Li-Bell, Adrian and Mothukuri, Mohith and Nair, Suraj and Pertsch, Karl and Ren, Allen Z. and Shi, Lucy Xiaoyang and Smith, Laura and Springenberg, Jost Tobias and Stachowicz, Kyle and Tanner, James and Vuong, Quan and Walke, Homer and Walling, Anna and Wang, Haohuan and Yu, Lili and Zhilinsky, Ury},
  journal       = {arXiv preprint arXiv:2504.16054},
  title         = {$\pi_{0.5}$: a Vision-Language-Action Model with Open-World Generalization},
  year          = {2025}
}

@article{keshishian2026parallel,
  author        = {Keshishian, Menoua and Mischler, Gavin and Thomas, Samuel and Kingsbury, Brian and Bickel, Stephan and Mehta, Ashesh D and Mesgarani, Nima},
  journal       = {Nature Machine Intelligence},
  number        = {2},
  pages         = {257--269},
  publisher     = {Nature Publishing Group UK London},
  title         = {Parallel hierarchical encoding of linguistic representations in the human auditory cortex and recurrent automatic speech recognition systems},
  volume        = {8},
  year          = {2026}
}

@inproceedings{kumar-etal-2022-gradient,
  address       = {Abu Dhabi, United Arab Emirates},
  author        = {Kumar, Sachin  and
Paria, Biswajit  and
Tsvetkov, Yulia},
  booktitle     = {Proceedings of the 2022 Conference on Empirical Methods in Natural Language Processing},
  doi           = {10.18653/v1/2022.emnlp-main.144},
  editor        = {Goldberg, Yoav  and
Kozareva, Zornitsa  and
Zhang, Yue},
  month         = {December},
  pages         = {2251--2277},
  publisher     = {Association for Computational Linguistics},
  title         = {Gradient-based Constrained Sampling from Language Models},
  year          = {2022}
}

@inproceedings{lewis2020retrieval,
  author        = {Lewis, Patrick and Perez, Ethan and Piktus, Aleksandra and Petroni, Fabio and Karpukhin, Vladimir and Goyal, Naman and K\"{u}ttler, Heinrich and Lewis, Mike and Yih, Wen-tau and Rockt\"{a}schel, Tim and Riedel, Sebastian and Kiela, Douwe},
  booktitle     = {Advances in Neural Information Processing Systems},
  editor        = {H. Larochelle and M. Ranzato and R. Hadsell and M.F. Balcan and H. Lin},
  pages         = {9459--9474},
  publisher     = {Curran Associates, Inc.},
  title         = {Retrieval-Augmented Generation for Knowledge-Intensive NLP Tasks},
  volume        = {33},
  year          = {2020}
}

@article{li2026environment,
  author        = {Li, Lin and Chen, Ziyang and Kan, Zhen},
  journal       = {IEEE Transactions on Automation Science and Engineering},
  publisher     = {IEEE},
  title         = {Environment-Driven and LLM-Guided Multi-Robot Task Inference and Allocation under Temporal Logic Specifications},
  year          = {2026}
}

@inproceedings{liu2023lang2ltl,
  author        = {Liu, Jason Xinyu and Yang, Ziyi and Idrees, Ifrah and Liang, Sam and Schornstein, Benjamin and Tellex, Stefanie and Shah, Ankit},
  booktitle     = {Proceedings of The 7th Conference on Robot Learning},
  editor        = {Tan, Jie and Toussaint, Marc and Darvish, Kourosh},
  month         = {06--09 Nov},
  pages         = {1084--1110},
  publisher     = {PMLR},
  series        = {Proceedings of Machine Learning Research},
  title         = {Grounding Complex Natural Language Commands for Temporal Tasks in Unseen Environments},
  volume        = {229},
  year          = {2023}
}

@inproceedings{liu2024lang2ltl,
  author        = {Liu, Jason Xinyu and Shah, Ankit and Konidaris, George and Tellex, Stefanie and Paulius, David},
  booktitle     = {2024 IEEE/RSJ International Conference on Intelligent Robots and Systems (IROS)},
  organization  = {IEEE},
  pages         = {2325--2332},
  title         = {Lang2ltl-2: Grounding spatiotemporal navigation commands using large language and vision-language models},
  year          = {2024}
}

@inproceedings{liu2024we,
  author        = {Liu, Michael Xieyang and Liu, Frederick and Fiannaca, Alexander J and Koo, Terry and Dixon, Lucas and Terry, Michael and Cai, Carrie J},
  booktitle     = {Extended Abstracts of the CHI Conference on Human Factors in Computing Systems},
  pages         = {1--9},
  title         = {" we need structured output": Towards user-centered constraints on large language model output},
  year          = {2024}
}

@article{liu2026hard,
  author        = {Liu, Yang and Zhou, Chuan and Chen, Yancheng and Zhang, Shuai and Lin, Xixun and Wang, Xiaoqing},
  journal       = {arXiv preprint arXiv:2602.01090},
  title         = {Hard Constraints Meet Soft Generation: Guaranteed Feasibility for LLM-based Combinatorial Optimization},
  year          = {2026}
}

@article{liu2026zero,
  author        = {Liu, Ruijia and Hou, Ancheng and Yu, Xiao and Yin, Xiang},
  journal       = {Advances in Neural Information Processing Systems},
  pages         = {130405--130442},
  title         = {Zero-shot trajectory planning for signal temporal logic tasks},
  volume        = {38},
  year          = {2026}
}

@inproceedings{loula2025syntactic,
  author        = {Loula, Jo{\~a}o and LeBrun, Benjamin and Du, Li and Lipkin, Ben and Pasti, Clemente and Grand, Gabriel and Liu, Tianyu and Emara, Yahya and Freedman, Marjorie and Eisner, Jason and others},
  booktitle     = {International Conference on Learning Representations},
  pages         = {64758--64791},
  title         = {Syntactic and semantic control of large language models via sequential monte carlo},
  volume        = {2025},
  year          = {2025}
}

@inproceedings{ma2025bridging,
  author        = {Ma, Zhi and Wen, Cheng and Su, Zhexin and Liang, Xiao and Tian, Cong and Qin, Shengchao and Yang, Mengfei},
  booktitle     = {2025 40th IEEE/ACM International Conference on Automated Software Engineering (ASE)},
  organization  = {IEEE},
  pages         = {1208--1220},
  title         = {Bridging natural language and formal specification--automated translation of software requirements to LTL via hierarchical semantics decomposition using LLMs},
  year          = {2025}
}

@article{ma2026survey,
  author        = {Ma, Yueen and Song, Zixing and Zhuang, Yuzheng and Hao, Jianye and King, Irwin},
  journal       = {IEEE Transactions on Neural Networks and Learning Systems},
  publisher     = {IEEE},
  title         = {A survey on vision--language--action models for embodied ai},
  year          = {2026}
}

@inproceedings{mavrogiannis2024cook2ltl,
  author        = {Mavrogiannis, Angelos and Mavrogiannis, Christoforos and Aloimonos, Yiannis},
  booktitle     = {2024 IEEE International Conference on Robotics and Automation (ICRA)},
  organization  = {IEEE},
  pages         = {17679--17686},
  title         = {Cook2ltl: Translating cooking recipes to ltl formulae using large language models},
  year          = {2024}
}

@inproceedings{mendoza2024translating,
  author        = {Mendoza, Daniel and Hahn, Christopher and Trippel, Caroline},
  booktitle     = {2024 Formal Methods in Computer-Aided Design (FMCAD)},
  organization  = {IEEE},
  pages         = {1--11},
  title         = {Translating natural language to temporal logics with large language models and model checkers},
  year          = {2024}
}

@inproceedings{mikolov2010recurrent,
  author        = {Mikolov, Tomas and Karafi{\'a}t, Martin and Burget, Lukas and Cernock{\`y}, Jan and Khudanpur, Sanjeev},
  booktitle     = {Interspeech},
  organization  = {Makuhari},
  pages         = {1045--1048},
  title         = {Recurrent neural network based language model.},
  volume        = {2},
  year          = {2010}
}

@article{nguyen2026thinking,
  author        = {Nguyen, Ngoc Trinh Hung and Silva, Alonso and Zumot, Laith and Tupikina, Liubov and Aghasaryan, Armen and Alam, Mehwish},
  journal       = {arXiv preprint arXiv:2601.07525},
  title         = {Thinking Before Constraining: A Unified Decoding Framework for Large Language Models},
  year          = {2026}
}

@inproceedings{oh2019planning,
  author        = {Oh, Yoonseon and Patel, Roma and Nguyen, Thao and Huang, Baichuan and Pavlick, Ellie and Tellex, Stefanie},
  booktitle     = {Robotics: Science and Systems XV},
  doi           = {10.15607/rss.2019.xv.059},
  pages         = {1--10},
  title         = {Planning with State Abstractions for Non-Markovian Task Specifications},
  year          = {2019}
}

@inproceedings{pakonen2016user,
  author        = {Pakonen, Antti and Pang, Cheng and Buzhinsky, Igor and Vyatkin, Valeriy},
  booktitle     = {2016 IEEE 21st International Conference on Emerging Technologies and Factory Automation (ETFA)},
  organization  = {IEEE},
  pages         = {1--8},
  title         = {User-friendly formal specification languages-conclusions drawn from industrial experience on model checking},
  year          = {2016}
}

@inproceedings{pan2023data,
  author        = {Pan, Jiayi and Chou, Glen and Berenson, Dmitry},
  booktitle     = {2023 IEEE International Conference on Robotics and Automation (ICRA)},
  organization  = {IEEE},
  pages         = {11554--11561},
  title         = {Data-Efficient Learning of Natural Language to Linear Temporal Logic Translators for Robot Task Specification},
  year          = {2023}
}

@inproceedings{parekh2025dicore,
  author        = {Parekh, Tanmay and Mehta, Kartik and Mehrabi, Ninareh and Chang, Kai-Wei and Peng, Nanyun},
  booktitle     = {Proceedings of the 2025 Conference on Empirical Methods in Natural Language Processing},
  pages         = {20571--20593},
  title         = {DiCoRe: Enhancing Zero-shot Event Detection via Divergent-Convergent LLM Reasoning},
  year          = {2025}
}

@inproceedings{patel2020grounding,
  author        = {Patel, Roma and Pavlick, Ellie and Tellex, Stefanie},
  booktitle     = {Robotics: Science and Systems},
  pages         = {1--12},
  title         = {Grounding Language to Non-Markovian Tasks with No Supervision of Task Specifications},
  volume        = {2020},
  year          = {2020}
}

@inproceedings{puig2018virtualhome,
  author        = {Puig, Xavier and Ra, Kevin and Boben, Marko and Li, Jiaman and Wang, Tingwu and Fidler, Sanja and Torralba, Antonio},
  booktitle     = {Proceedings of the IEEE conference on computer vision and pattern recognition},
  pages         = {8494--8502},
  title         = {Virtualhome: Simulating household activities via programs},
  year          = {2018}
}

@inproceedings{quansah2026neuronl2ltl,
  author        = {Quansah, Paapa Kwesi and Bonnah, Ernest},
  booktitle     = {International Conference on Formal Techniques for Distributed Objects, Components, and Systems},
  organization  = {Springer},
  pages         = {56--73},
  title         = {NeuroNL2LTL: A Neurosymbolic Framework for Natural Language Translation of Linear Temporal Logic},
  year          = {2026}
}

@inproceedings{quartey2025verifiably,
  author        = {Quartey, Benedict and Rosen, Eric and Tellex, Stefanie and Konidaris, George},
  booktitle     = {2025 IEEE International Conference on Robotics and Automation (ICRA)},
  organization  = {IEEE},
  pages         = {1--8},
  title         = {Verifiably following complex robot instructions with foundation models},
  year          = {2025}
}

@inproceedings{rabiei2025ltlcodegen,
  author        = {Rabiei, Behrad and AR, Mahesh Kumar and Dai, Zhirui and Pilla, Surya LSR and Dong, Qiyue and Atanasov, Nikolay},
  booktitle     = {2025 IEEE/RSJ International Conference on Intelligent Robots and Systems (IROS)},
  organization  = {IEEE},
  pages         = {19240--19247},
  title         = {Ltlcodegen: Code generation of syntactically correct temporal logic for robot task planning},
  year          = {2025}
}

@inproceedings{rajpurkar2016squad,
  author        = {Rajpurkar, Pranav and Zhang, Jian and Lopyrev, Konstantin and Liang, Percy},
  booktitle     = {Proceedings of the 2016 Conference on Empirical Methods in Natural Language Processing},
  pages         = {2383--2392},
  title         = {SQuAD: 100,000+ Questions for Machine Comprehension of Text},
  year          = {2016}
}

@inproceedings{raman2014model,
  author        = {Raman, Vasumathi and Donz{\'e}, Alexandre and Maasoumy, Mehdi and Murray, Richard M and Sangiovanni-Vincentelli, Alberto and Seshia, Sanjit A},
  booktitle     = {53rd IEEE Conference on Decision and Control},
  organization  = {IEEE},
  pages         = {81--87},
  title         = {Model predictive control with signal temporal logic specifications},
  year          = {2014}
}

@inproceedings{rankin2021robotic,
  author        = {Rankin, Ian C and McCammon, Seth and Hollinger, Geoffrey A},
  booktitle     = {2021 IEEE International Conference on Robotics and Automation (ICRA)},
  organization  = {IEEE},
  pages         = {4882--4888},
  title         = {Robotic information gathering using semantic language instructions},
  year          = {2021}
}

@article{rayner1975perceptual,
  author        = {Rayner, Keith},
  journal       = {Cognitive psychology},
  number        = {1},
  pages         = {65--81},
  publisher     = {Elsevier},
  title         = {The perceptual span and peripheral cues in reading},
  volume        = {7},
  year          = {1975}
}

@inproceedings{santos2025updating,
  author        = {Santos, Leonardo and Li, Zirui and Peters, Lasse and Bansal, Somil and Bajcsy, Andrea},
  booktitle     = {2025 IEEE International Conference on Robotics and Automation (ICRA)},
  organization  = {IEEE},
  pages         = {7778--7785},
  title         = {Updating robot safety representations online from natural language feedback},
  year          = {2025}
}

@incollection{schlor2006using,
  author        = {Schl{\"o}r, Rainer and Josko, Bernhard and Werth, Dieter},
  booktitle     = {Services and Visualization Towards User-Friendly Design: ACoS'98, VISUAL'98, AIN'97 Selected Papers},
  pages         = {208--221},
  publisher     = {Springer},
  title         = {Using a visual formalism for design verification in industrial environments},
  year          = {2006}
}

@misc{schulman2017proximal,
  archiveprefix = {arXiv},
  author        = {John Schulman and Filip Wolski and Prafulla Dhariwal and Alec Radford and Oleg Klimov},
  eprint        = {1707.06347},
  primaryclass  = {cs.LG},
  title         = {Proximal Policy Optimization Algorithms},
  year          = {2017}
}

@inproceedings{shi2024autodsl,
  author        = {Shi, Yu-Zhe and Hou, Haofei and Bi, Zhangqian and Meng, Fanxu and Wei, Xiang and Ruan, Lecheng and Wang, Qining},
  booktitle     = {Proceedings of the 62nd Annual Meeting of the Association for Computational Linguistics (Volume 1: Long Papers)},
  pages         = {12177--12214},
  title         = {AutoDSL: Automated domain-specific language design for structural representation of procedures with constraints},
  year          = {2024}
}

@inproceedings{sun2025detection,
  author        = {Zhongxiang Sun and Qipeng Wang and Haoyu Wang and Xiao Zhang and Jun Xu},
  booktitle     = {Socially Responsible and Trustworthy Foundation Models at NeurIPS 2025},
  pages         = {1--25},
  title         = {Detection and Mitigation of Hallucination in Large Reasoning Models: A Mechanistic Perspective},
  year          = {2025}
}

@inproceedings{tellex2011understanding,
  author        = {Tellex, Stefanie and Kollar, Thomas and Dickerson, Steven and Walter, Matthew and Banerjee, Ashis and Teller, Seth and Roy, Nicholas},
  booktitle     = {Proceedings of the AAAI conference on artificial intelligence},
  pages         = {1507--1514},
  title         = {Understanding natural language commands for robotic navigation and mobile manipulation},
  volume        = {25},
  year          = {2011}
}

@article{verkerk2026enduring,
  author        = {Verkerk, Annemarie and Shcherbakova, Olena and Haynie, Hannah J and Skirg{\aa}rd, Hedvig and Rzymski, Christoph and Atkinson, Quentin D and Greenhill, Simon J and Gray, Russell D},
  journal       = {Nature human behaviour},
  number        = {1},
  pages         = {126--136},
  publisher     = {Nature Publishing Group UK London},
  title         = {Enduring constraints on grammar revealed by Bayesian spatiophylogenetic analyses},
  volume        = {10},
  year          = {2026}
}

@inproceedings{wang2022self,
  author        = {Xuezhi Wang and Jason Wei and Dale Schuurmans and Quoc V Le and Ed H. Chi and Sharan Narang and Aakanksha Chowdhery and Denny Zhou},
  booktitle     = {The Eleventh International Conference on Learning Representations},
  pages         = {1--24},
  title         = {Self-Consistency Improves Chain of Thought Reasoning in Language Models},
  year          = {2023}
}

@article{wang2025conformalnl2ltl,
  author        = {Wang, Jun and Sundarsingh, David Smith and Deshmukh, Jyotirmoy V and Kantaros, Yiannis},
  journal       = {arXiv preprint arXiv:2504.21022},
  title         = {Conformalnl2ltl: Translating natural language instructions into temporal logic formulas with conformal correctness guarantees},
  year          = {2025}
}

@article{wang2026logicflow,
  author        = {Wang, Shaochen and Wu, Qilin and Xia, Yanjie},
  journal       = {IEEE Robotics and Automation Letters},
  publisher     = {IEEE},
  title         = {LogicFlow: Amortizing Signal Temporal Logic into Continuous Vector Fields for Safe Driving},
  year          = {2026}
}

@inproceedings{wei2022chain,
  author        = {Wei, Jason and Wang, Xuezhi and Schuurmans, Dale and Bosma, Maarten and ichter, brian and Xia, Fei and Chi, Ed and Le, Quoc V and Zhou, Denny},
  booktitle     = {Advances in Neural Information Processing Systems},
  editor        = {S. Koyejo and S. Mohamed and A. Agarwal and D. Belgrave and K. Cho and A. Oh},
  pages         = {24824--24837},
  publisher     = {Curran Associates, Inc.},
  title         = {Chain-of-Thought Prompting Elicits Reasoning in Large Language Models},
  volume        = {35},
  year          = {2022}
}

@article{willard2023efficient,
  author        = {Willard, Brandon T and Louf, R{\'e}mi},
  journal       = {arXiv preprint arXiv:2307.09702},
  title         = {Efficient guided generation for large language models},
  year          = {2023}
}

@inproceedings{wu2025selp,
  author        = {Wu, Yi and Xiong, Zikang and Hu, Yiran and Iyengar, Shreyash S and Jiang, Nan and Bera, Aniket and Tan, Lin and Jagannathan, Suresh},
  booktitle     = {2025 IEEE International Conference on Robotics and Automation (ICRA)},
  organization  = {IEEE},
  pages         = {2599--2605},
  title         = {SELP: Generating safe and efficient task plans for robot agents with large language models},
  year          = {2025}
}

@article{xu2025toward,
  author        = {Xu, Fengli and Hao, Qianyue and Shao, Chenyang and Zong, Zefang and Li, Yu and Wang, Jingwei and Zhang, Yunke and Wang, Jingyi and Lan, Xiaochong and Gong, Jiahui and Ouyang, Tianjian and Meng, Fanjin and Yan, Yuwei and Yang, Qinglong and Song, Yiwen and Ren, Sijian and Hu, Xinyuan and Feng, Jie and Gao, Chen and Li, Yong},
  journal       = {Patterns},
  number        = {10},
  publisher     = {Elsevier},
  title         = {Toward large reasoning models: A survey of reinforced reasoning with large language models},
  volume        = {6},
  year          = {2025}
}

@article{yao2023tree,
  author        = {Yao, Shunyu and Yu, Dian and Zhao, Jeffrey and Shafran, Izhak and Griffiths, Tom and Cao, Yuan and Narasimhan, Karthik},
  journal       = {Advances in neural information processing systems},
  pages         = {11809--11822},
  title         = {Tree of thoughts: Deliberate problem solving with large language models},
  volume        = {36},
  year          = {2023}
}

@article{yao2025reasoning,
  author        = {Yao, Zijun and Liu, Yantao and Chen, Yanxu and Chen, Jianhui and Fang, Junfeng and Hou, Lei and Li, Juanzi and Chua, Tat-Seng},
  journal       = {arXiv preprint arXiv:2505.23646},
  title         = {Are reasoning models more prone to hallucination?},
  year          = {2025}
}

@misc{zeng2025simplerlzooinvestigatingtamingzero,
  archiveprefix = {arXiv},
  author        = {Weihao Zeng and Yuzhen Huang and Qian Liu and Wei Liu and Keqing He and Zejun Ma and Junxian He},
  eprint        = {2503.18892},
  primaryclass  = {cs.LG},
  title         = {SimpleRL-Zoo: Investigating and Taming Zero Reinforcement Learning for Open Base Models in the Wild},
  year          = {2025}
}

@inproceedings{zheng2024llamafactory,
  address       = {Bangkok, Thailand},
  author        = {Zheng, Yaowei  and
Zhang, Richong  and
Zhang, Junhao  and
Ye, Yanhan  and
Luo, Zheyan},
  booktitle     = {Proceedings of the 62nd Annual Meeting of the Association for Computational Linguistics (Volume 3: System Demonstrations)},
  doi           = {10.18653/v1/2024.acl-demos.38},
  editor        = {Cao, Yixin  and
Feng, Yang  and
Xiong, Deyi},
  month         = {August},
  pages         = {400--410},
  publisher     = {Association for Computational Linguistics},
  title         = {{L}lama{F}actory: Unified Efficient Fine-Tuning of 100+ Language Models},
  year          = {2024}
}

@inproceedings{zhou2026policy,
  author        = {Zhou, Jianpeng and Hu, Qisheng and Wang, Jiahai and Wang, Wenya},
  booktitle     = {Findings of the Association for Computational Linguistics: ACL 2026},
  pages         = {40740--40765},
  title         = {Policy-Guided Stepwise Action Planning for Controllable LLM Reasoning},
  year          = {2026}
}

\end{document}